\documentclass[10pt,journal,compsoc]{IEEEtran}
\IEEEoverridecommandlockouts
\usepackage{cite}
\usepackage{amsmath,amssymb,amsfonts}
\usepackage{algorithmic}
\usepackage{graphicx}
\usepackage{textcomp}
\usepackage{xcolor}
\usepackage{amsthm,amsmath,amssymb}
\usepackage{booktabs}
\usepackage{subfigure}

\newtheorem*{claim}{Claim}

\usepackage{mathrsfs}
\def\BibTeX{{\rm B\kern-.05em{\sc i\kern-.025em b}\kern-.08em
    T\kern-.1667em\lower.7ex\hbox{E}\kern-.125emX}}

\begin{document}

\title{Making DeepFakes more spurious: evading deep face forgery detection via trace removal attack}
\author{Chi~Liu, 
        Huajie~Chen,
        Tianqing~Zhu*,
        Jun~Zhang,
        Wanlei~Zhou
\thanks{*Tianqing Zhu is the corresponding author.}
\thanks{Chi Liu, Huajie Chen and Tianqing Zhu are with the Centre for Cyber Security and Privacy and the School of Computer Science, University of Technology Sydney, Sydney, NSW 2007, Australia (e-mail: chi.liu@student.uts.edu.au; huajie.chen@student.uts.edu.au; tianqing.zhu@uts.edu.au; ).}
\thanks{Jun Zhang is with the Cybersecurity Lab, Swinburne University of Technology, Melbourne, VIC 2008, Australia (e-mail: junzhang@swin.edu.au).}
\thanks{Wanlei Zhou is with the Institute of Data Science, City University of Macau, Macao SAR, China (e-mail: wlzhou@cityu.edu.mo).}}

\markboth{Submitted to IEEE,~Vol.~xx, No.~xx, Mar~2022}%
{Shell \MakeLowercase{\textit{et al.}}: Bare Demo of IEEEtran.cls for IEEE Journals}

\maketitle

\begin{abstract}
DeepFakes are raising significant social concerns. Although various DeepFake detectors have been developed as forensic countermeasures, these detectors are still vulnerable to attacks. Recently, a few attacks, principally adversarial attacks, have succeeded in cloaking DeepFake images to evade detection. However, these attacks have typical detector-specific designs, which require prior knowledge about the detector, leading to poor transferability. Moreover, these attacks only consider simple security scenarios. Less is known about how effective they are in high-level scenarios where either the detector’s defensive capability or the attacker’s knowledge varies. In this paper, we aim to solve the above challenges with presenting a novel attack pattern for DeepFake anti-forensics, namely, the trace removal attack. Instead of investigating the detector side, this trace removal attack looks into the original DeepFake creation pipeline, attempting to remove all detectable natural DeepFake traces to render the fake images more “authentic”. This detector-agnostic design benefits the attack to be effective against arbitrary or even unknown detectors. To implement this attack, first, we perform an in-depth DeepFake trace discovery, which identifies three discernible traces: spatial anomalies, spectral disparities, and noise fingerprints. Then a trace removal network (TR-Net) is proposed based on an adversarial learning framework involving one generator and multiple discriminators. Each discriminator is responsible for one individual trace representation to avoid cross-trace interference. These multiple discriminators are arranged in parallel, which prompts the generator to remove various traces simultaneously. To evaluate the efficacy of the attack, we crafted heterogeneous security scenarios where the detectors were embedded with different levels of defense and the attackers' background knowledge of data varies. The experimental results show that the proposed attack can significantly compromise the detection accuracy of six state-of-the-art DeepFake detectors while causing only a negligible loss in visual quality to the original DeepFake samples.

\end{abstract}


\section{Introduction}
Along with the recent progress in automated digital face manipulation techniques based on deep learning, deep face forgeries, also known as DeepFakes, are raising serious social concerns for information security \cite{westerlund2019emergence}. Accordingly, the research community is dedicated to developing forensic countermeasures against DeepFakes, and many DeepFake detectors have been developed that can successfully distinguish DeepFake images from real ones \cite{tolosana2020deepfakes}. However, the robustness of these detectors to malicious attacks is still in the early stages. To further understand the vulnerability of DeepFake detectors, researchers have and must continue to engage in anti-forensics against DeepFake detection \cite{hussain2021adversarial, neekhara2021adversarial, carlini2020evading, gandhi2020adversarial, fan2021deepfake, liao2021imperceptible, wang2021perception, huang2020fakeretouch, ding2021anti, neves2020ganprintr, huang2020fakepolisher}. Each novel anti-forensic attack exposed can help us to analyze these detectors more comprehensively.

Most existing attacks are based on adversarial attacks which embed imperceptible adversarial perturbations into DeepFake samples to fool machine learning-based detectors \cite{hussain2021adversarial, neekhara2021adversarial, carlini2020evading, gandhi2020adversarial, fan2021deepfake, liao2021imperceptible, wang2021perception}. The development of this type of attack relies on the background knowledge of detectors, such as the queried outputs and the detector’s parameters. Even in a universal black-box attack scenario, information from surrogate detectors is always needed to imitate the behavior of the target detector. These \textit{detector-specific} designs lead to poor transferability and a lack of stability across different detectors or unknown detectors \cite{barni2019transferability, zhao2020effect}. Other attacks emerging in this field generally require reconstructing the DeepFake samples to modify the distribution of features-of-interest of the target detector to evade detection \cite{huang2020fakeretouch, ding2021anti, neves2020ganprintr, huang2020fakepolisher}. This is also a detector-specific design, which means that these attacks are less transferable to detectors interested in different forgery features. Moreover, these attacks only pay attention to a single type of feature. Their efficacy may deteriorate significantly against advanced detectors that operate on hybrid features. 

\begin{figure}[]
    \centering
     \includegraphics[width=0.45\textwidth]{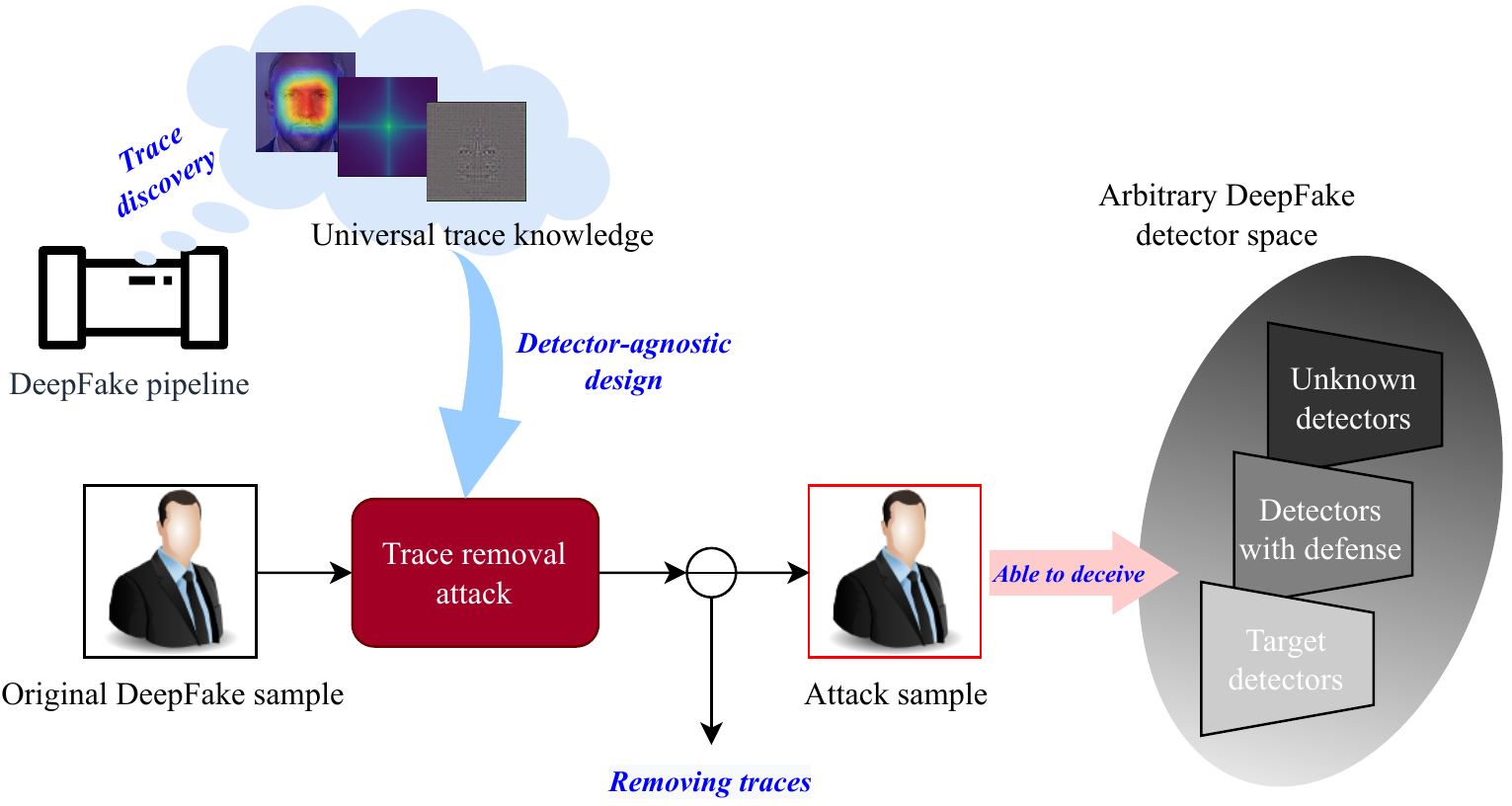}
     \caption{The proposed trace removal attack utilizes the universal trace knowledge distilled from the common DeepFake pipeline instead of detector-specific knowledge. Thus it is detector-agnostic and can transfer across arbitrary black-box detectors.}
     \label{fig:design}
 \end{figure}

Another weakness of these attacks is that their studies tend to oversimplify the security scenarios. On the one hand, the attacks are often implemented and evaluated with ideal assumptions, e.g., “the attacker has unlimited access to the target detector” (or at least the surrogates) or “the attacker has all required background knowledge of the training data”. Second, the target detectors are often assumed to be as naked as possible, while some common and easy-to-implement defenses are left out of consideration.

In this paper, we propose a novel attack pattern for DeepFake anti-forensics, called the \textit{trace removal attack}, that addresses the above weaknesses. Unlike the detector-specific designs, we offer a novel \textit{detector-agnostic} perspective. As shown in Figure \ref{fig:design}, we pay full attention to the original pipeline of DeepFake image creation, identifying the discernible manufacturing traces in the DeepFake images. The DeepFake images are then refined by removing all these traces, resulting in images (i.e., attack samples) that are able to bypass any arbitrary detector. Our attack requires little knowledge of the target detector, operating exclusively on the DeepFake images without any additional interactions with the target detector. In contrast to adding extra adversarial noise or modifying the feature distribution, removing the intrinsic detectable traces makes the DeepFake images essentially much closer to the real ones, i.e., the DeepFake images become more natural and perceptually “authentic”. In this sense, the proposed method can be seen as a universal black-box attack.

To implement the trace removal attack, the first step is to conduct an empirical trace discovery to thoroughly investigate what discernible manufacturing traces are naturally maintained in DeepFake images. An adversarial learning-based trace removal network  (TR-Net) then removes the traces found. However, unlike a normal adversarial learning network with one generator and one discriminator, TR-Net contains a single generator and multiple discriminators, where each discriminator is responsible for distinguishing one particular type of trace. This “one-versus-multiple” structure can prompt the generator to reconstruct DeepFake images by removing all possible traces synchronously. Considering the identified traces could exist in different signal domains, using multiple discriminators allows the representation of these traces to be effectively decoupled. We constructed several heterogeneous threat scenarios to assess the efficacy of the attack, where the detectors are reinforced through various defensive strategies, and the attackers have different data background knowledge. We then evaluated the attack on a wide range of representative detectors to ensure this detector-agnostic attack is truly universal and transferable. 

Our contributions are as follows:
\begin{itemize}
    \item We performed an in-depth DeepFake trace discovery, identifying three universal traces responsible for DeepFake images’ tractability.   
    \item We propose a novel attack concept against DeepFake detectors, namely, the trace removal attack. Benefiting from a detector-agnostic design, our attack can defeat arbitrary unknown detectors and detectors equipped with defenses. The attack is implemented via a “one-versus-multiple” adversarial learning network that erases all traces synchronously. 
    \item The attack is tested in heterogeneous threat scenarios, where the detector’s defensive capability ranges from weak to strong and the attacker’s data knowledge is limited. Further, performance is evaluated on a wide range of detectors, and a dataset is developed covering all typical DeepFake types to benchmark our evaluation.
\end{itemize}

\section{Preliminary and related work}

\subsection{Generative adversarial net}

Generative adversarial net (GAN) \cite{goodfellow2014generative} and its variants are a typical type of deep generative models that serve as a key technique for generating DeepFakes \cite{tolosana2020deepfakes}. A GAN normally involves a generator $G$ and a discriminator $D$ in an adversarial learning framework where the aim is to train $G$ to generate synthetic images within the target distribution $p_{t}(x)$. The process can be described as: 

\footnotesize
\begin{equation}
\min _{G} \max _{D} T(G, D)=\mathbb{E}_{p_{t}(\boldsymbol{x})}[\log D(\boldsymbol{x})]+\mathbb{E}_{p_{\boldsymbol{z}}(\boldsymbol{z})}[\log (1-D(G(\boldsymbol{z})))].
\end{equation}
\normalsize

If an additional control is imposed over the modes of the data to be generated, such as adding an attribute label $y$ as prior guidance, a conditional GAN will be created then.

\subsection{DeepFake}
\label{sec:deepfake}
DeepFakes are generated in roughly one of three ways: face synthesis, facial attribute editing, or face replacement \cite{tolosana2020deepfakes}. Face synthesis means to create an entire non-existent face from random noise with an unconditional GAN, such as ProGAN \cite{karras2018progressive} and StyleGAN \cite{karras2019style}. With facial attribute editing, an image’s attributes are altered. Either the appearance attributes (e.g., hair color, makeup, skin color, etc.) or the soft biometric attributes (e.g., identity, gender, age, etc.) can be modified. Conditional GANs, such as StarGAN \cite{choi2018stargan} and STGAN \cite{liu2019stgan} are widely employed for such tasks. Here, the target attribute serves as the extra label $y$ in training a conditional GAN. Face replacement swaps the face of a target image with that of a source image. Factors that need to be considered include the alignment of the face in terms of size, pose, and direction. A deep rendering process then ensures the resulting image looks natural and seamless. In addition, these methods can be combined to produce high-level DeepFakes, like high-fidelity facial reenactments for fake videos.

\subsection{DeepFake Forensics}
\label{sec:detectors}

DeepFake forensics is a discipline devoted to detect whether or not an image is a DeepFake image. There are three main categories of detectors: spatial detectors, frequency detectors, and fingerprint detectors.

\textbf{Spatial detectors} learn discriminative features directly from the spatial information of the images, i.e., the pixel inputs. The available features range from low-level cues to deep representative features. For example, some researchers classify DeepFakes versus real images based on disparities in their color components \cite{matern2019exploiting, li2020identification, chen2021robust}, global image textures \cite{liu2020global}, or the consistency of the facial context \cite{nirkin2021deepfake}. Others train more generalized detectors by simulating the shared visual artifacts in DeepFakes led by face manipulation \cite{li2019exposing} or blending \cite{nirkin2021deepfake}. Chai et al. \cite{chai2020makes} investigated the semantic properties that make DeepFakes detectable at the image patch level. There are also several studies \cite{rossler2019faceforensics++, wang2020cnn, das2021towards} that investigate the power of deep convolutional neural networks (CNNs), such as Xception \cite{chollet2017xception} and ResNet \cite{he2016deep}, in DeepFake detection from a data-driven perspective.

\textbf{Frequency-based detectors} solve DeepFake detection using frequency cues. These detectors normally mine distinctive features from the image spectra via statistical analysis or machine learning. For instance, Durall et al. \cite{durall2020watch}, Dzanic et al. \cite{dzanic2020fourier} and Frank et al. \cite{frank2020leveraging} determined that there were Fourier spectrum discrepancies between CNN-generated images and real images which could be efficiently captured by a shallow machine learning classifier. Liu et al. \cite{liu2021spatial} highlight that these discrepancies are more significant in phase spectra than in amplitude spectra, which is helpful for DeepFake detection. Zhang et al. \cite{zhang2019detecting} identify unknown DeepFakes by detecting and simulating general spectrum artifacts. Qian et al. \cite{qian2020thinking} mine frequency-aware clues from both frequency-aware decomposed image components and local frequency statistics. Some more recent studies \cite{chen2021local, luo2021generalizing} have put forward hybrid frameworks that combine both spatial and frequency information to generalize DeepFake detection.

\textbf{Fingerprint detectors.} Similar to cameras leaving their device fingerprints on a photo, researchers have discovered that GANs will leave unique and stable fingerprints in their generated images, which can be exploited to differentiate GAN-generated DeepFakes from real images. For example, Marra et al. \cite{marra2019gans} estimate GAN fingerprints as average noise residuals for GAN-generated image forensics. Yu et al. \cite{yu2019attributing} use a CNN to learn GAN fingerprint representations as a way of attributing DeepFakes to their source GANs. Yang et al. \cite{yang2021learning} devised a method of disentangling content-irrelevant fingerprints from GAN-generated images to determine DeepFakes. Yu et al. \cite{yu2021artificial} proposed introducing artificial fingerprints into the source models that produce DeepFakes for proactive and sustainable detection.

\subsection{Anti-forensics for DeepFakes}
\textbf{Adversarial attack:} Since most forgery detectors are machine learning models, adversarial attacks, as a typical type of attack against machine learning classifiers, have become a common anti-forensic choice for DeepFakes. A successful adversarial attack requires embedding imperceptible noise perturbations into the fake sample, which deceives the detector into classifying the image as “real”. Several classic adversarial attack methods, including Fast Gradient Sign Method (FGSM) \cite{goodfellow2014explaining}, iterative FGSM \cite{kurakin2018adversarial}, Carlini and Wagner $l_2$-norm Attack \cite{carlini2017towards}, DeepFool \cite{moosavi2016deepfool} and Projected Gradient Descent (PGD) \cite{madry2017towards}, are explored to expose the vulnerability of DeepFake detectors in both white- and black-box scenarios \cite{hussain2021adversarial, neekhara2021adversarial, carlini2020evading, gandhi2020adversarial, fan2021deepfake, barni2019transferability, zhao2020effect}. Liao et al. \cite{liao2021imperceptible} improved on the efficiency of these attacks by adding perturbations to key regions of the DeepFakes instead of across the entire image. Wang et al. \cite{wang2021perception} point out that adding adversarial noise to a transformed color space will ameliorate the perceptual degradation in a perturbed DeepFake image. Huang et al. \cite{huang2020fakeretouch} proposed an adversarial-noise-guided filtering method that retouches DeepFake images to evade detection.

\textbf{Reconstruction-based attack:} Very recently, some other attacks were proposed based on reconstructing DeepFake samples instead of injecting external noise to the samples for evading forensic detection. Ding et al. \cite{ding2021anti} proposed an adversarial learning method that re-synthesizes face-swapping images with narrowing down the distribution gap between real and fake faces. Huang et al. \cite{huang2020fakepolisher} proposed FakePolisher, a shallow reconstruction model based on dictionary learning. The model projects DeepFake images onto the manifold subspace learned from real images to reduce spectral discrepancies between the two. Neves et al. \cite{neves2020ganprintr} proposed GANprintR, a deep convolutional autoencoder that learns the reconstruction-related representation from natural images. Then the potential traceable fingerprints in the DeepFake image can be removed by feeding the image to the learned autoencoder.


\section{Empirical DeepFake trace discovery}
\label{sec:DTD}
In this section, we investigate the original process that creates DeepFakes to provide insights into the root causes that make DeepFakes detectable. In this way, the universal forgery traces can be identified empirically. 

\subsection{The DeepFake pipeline}

\begin{figure}[]
    \centering
    \includegraphics[width=0.4\textwidth]{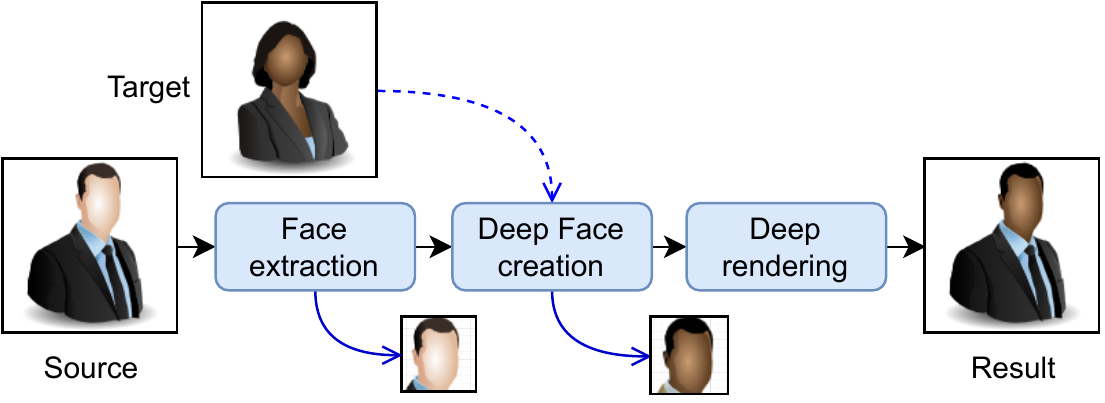}
    \caption{The fundamental DeepFake pipeline.}
    \label{fig:df_pipeline}
\end{figure}
The top-level design of a DeepFake generator may vary, but underneath there is a common pipeline for producing DeepFake images that consists of three core stages: face extraction, fake face creation, and deep rendering, as shown in Figure \ref{fig:df_pipeline}. In the first stage, the facial region is localized and extracted from the source image. This process can be accomplished without a GAN. Next, a fake face is generated by a specific GAN or a model template according to the target face. In most cases, creating a fake face is conditioned upon some knowledge of the target face, such as the identity or attributes. In the last stage, the generated face is aligned to the source face and composed back onto the source image. Usually, a GAN or some post-processing operations are employed to render the final image to make it more natural.

This pipeline is applicable to all the aforementioned methods of generating a DeepFake, either partially or entirely. For example, end-to-end facial synthesis and attribute editing with DeepFake models are \textit{de facto} productions from the second stage of the pipeline, while more sophisticated DeepFake models are equipped with a deep rendering process at the end of the pipeline.

\subsection{What make DeepFakes detectable?}
\subsubsection{Model traces in fake face creation}
\label{sec:trace}
\ 
\newline
\indent \textbf{Spatial anomalies.} DeepFakes rely on deep generative models, such as GANs, to synthesize faces from real face images. Ideally, the generated faces should be visually indistinguishable from real ones. However, due to some practical limitations e.g., with the dataset or the model’s capability, the fake faces may be imperfect which may show spatial abnormalities. Although the latest GANs have seen a significant improvement in visual quality over their predecessors, some subtle unnatural traces such as inconsistencies in texture or contextual discrepancies can still occur \cite{matern2019exploiting, li2020identification, chen2021robust, liu2020global, nirkin2021deepfake}. Spatial anomalies can also be found in faces generated by model templates. This can be a result of manufacturing failures during the template alignment or rendering \cite{li2019exposing, li2020face}. 

\begin{figure}[]
    \centering
    \includegraphics[width=0.44\textwidth]{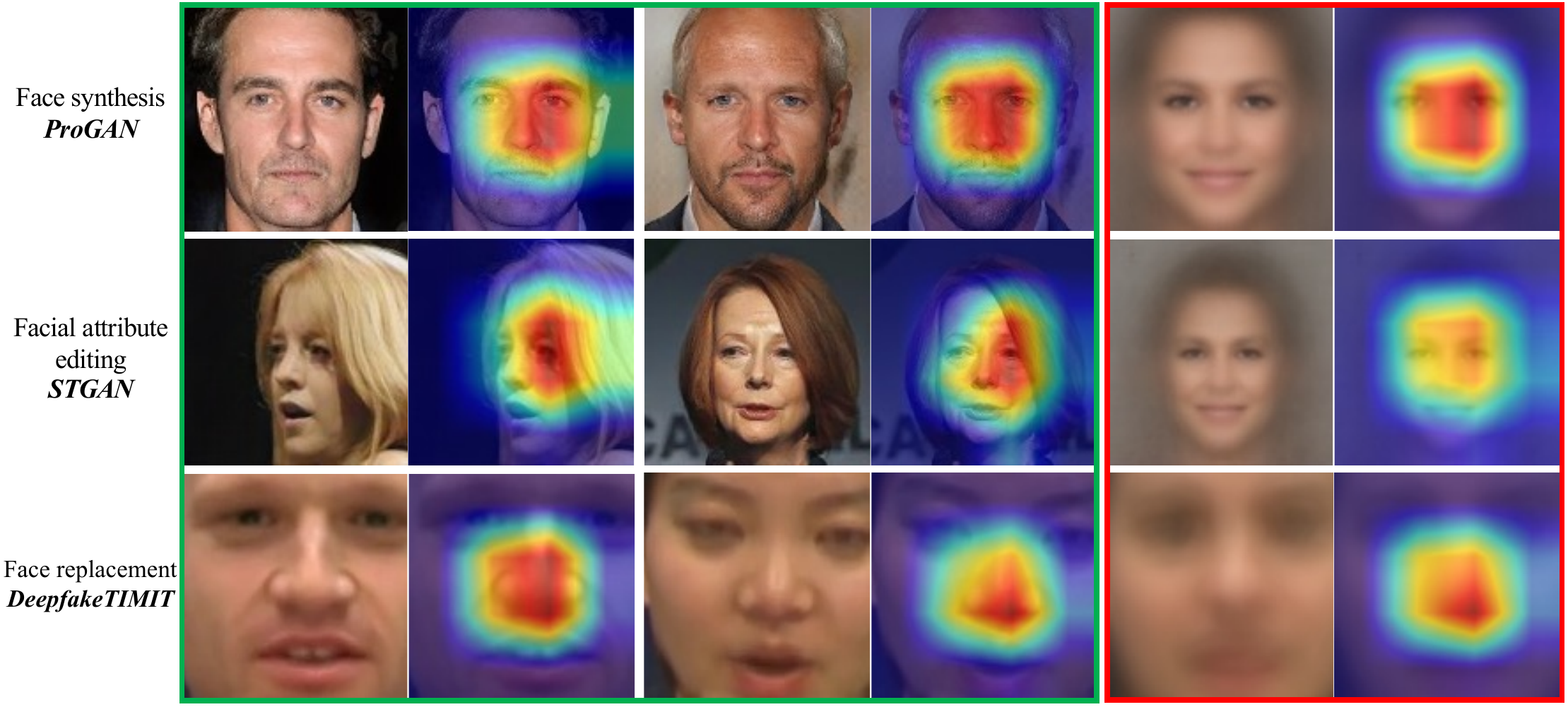}
    \caption{Spatial anomalies revealed by the spatial attention maps of a Xception detector. For each DeepFake type, the green box shows two random DeepFake examples and the red box shows the average results of $2000$ samples.}
    \label{fig:spatial}
\end{figure}

Notably, since the subtle spatial anomalies may be imperceptible to humans but can be captured by machines, we demonstrate their existence and spatial distributions in the RGB color space with the spatial attention map (SAM) of a toy Xception detector. Grad-CAM \cite{selvaraju2017grad} is used to calculate the SAMs regarding different DeepFake types (Details of these DeepFake types and the Xception detector is introduced in Section \ref{sec:experiment}). As shown in Figure \ref{fig:spatial}, there are evident detectable traces in the RGB space, and their distributions exhibit certain stable semantic-dependencies: the ProGAN and STGAN’s anomalies are around the middle-right face region while the DeepfakeTIMIT images expose traces concentrated in the nose area. These results can be seen more clearly in the averaged faces.

\textbf{Spectral disparity.}  Detectable traces can also be revealed in the frequency domain. This is because that CNN-based generative models, typically GANs, will create disparities in the spectra of the generated images. Some past studies attribute this phenomenon to the transposed convolution operation – a widely-used upsampling unit in CNN-based generative models for increasing feature dimensionality \cite{durall2020watch, chandrasegaran2021closer, zhang2019detecting, frank2020leveraging, liu2021spatial}. 


\begin{claim}[1]
The transposed convolution operation in upsampling layers leads to quasi-periodic high-frequency artifacts in the resulting feature maps. (The proof is in the supplement.) 
\end{claim}

\begin{figure}[]
    \centering
    \includegraphics[width=0.3\textwidth]{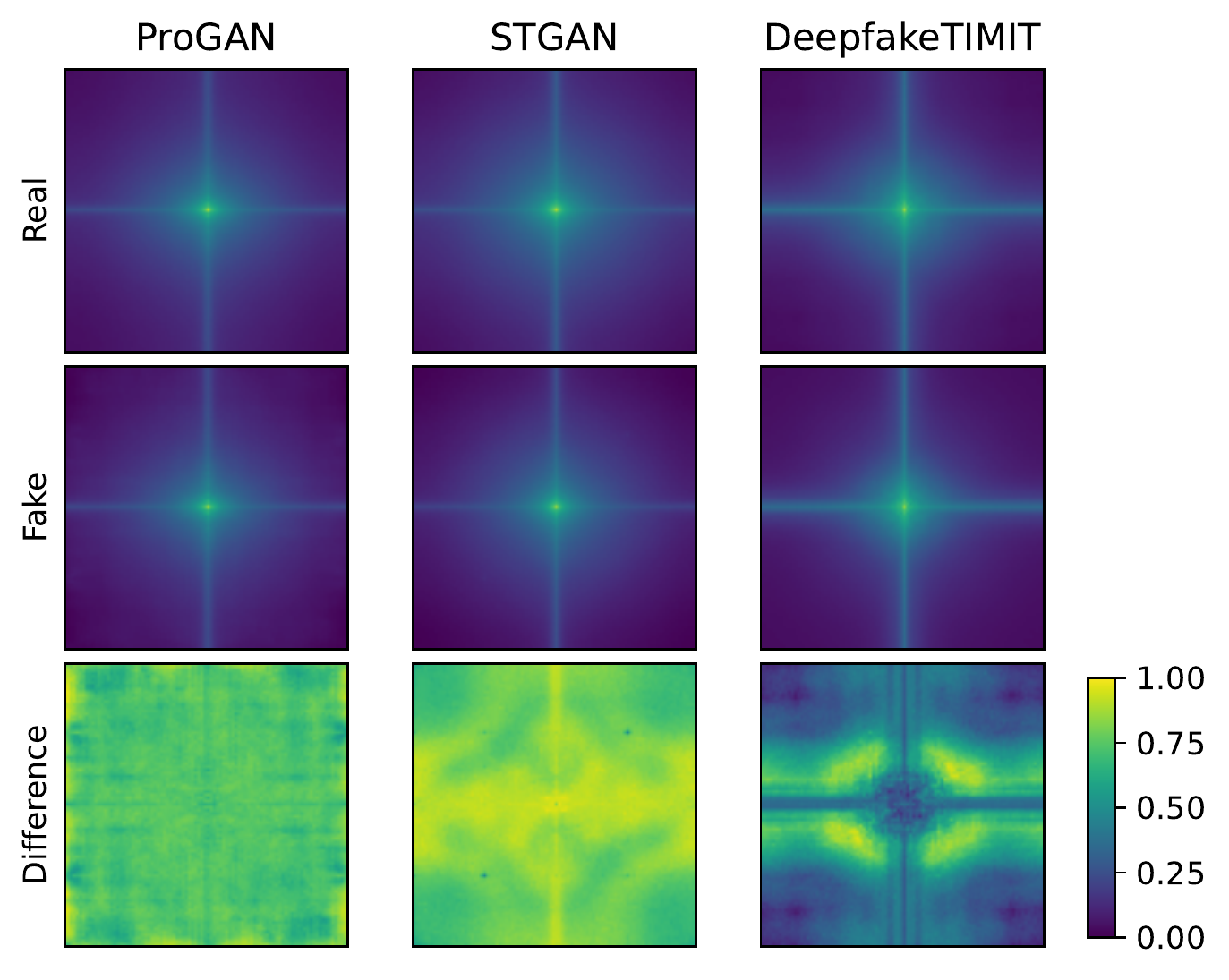}
    \caption{The averaged spectra of both the real images and the corresponding fake images of different DeepFake types. Each spectrum is averaged on $2000$ samples. The last row shows the differences between the spectra of the real and fake images.}
    \label{fig:spectral}
\end{figure}

An illustration of the disparity between the averaged spectra of real and fake images is provided in Figure \ref{fig:spectral}. All three types of DeepFake images have significant differences from the real ones. The disparity patterns in the DeepfakeTIMIT images are not similar to the other two because face replacement involves post-processing procedures that further change the spectral distribution.

\subsubsection{Model traces in deep rendering}
\ 
\newline
\indent \textbf{Noise fingerprint.} A DeepFake may retain two types of manufacturing fingerprints through the deep rendering phase, one being the GAN fingerprint, the other one the post-processing fingerprint. The deep rendering phase usually involves a GAN-based rendering model and some post-processing operations, such as landmark alignment, color correction, splicing, and blending. It has been pointed out that GANs maintain unique and stable fingerprints in their generated images. Likewise, post-processing operations will also introduce fingerprints, due to the characteristic discrepancies in the noise space brought about through tampering the regions.

\begin{figure}[]
    \centering
    \includegraphics[width=0.3\textwidth]{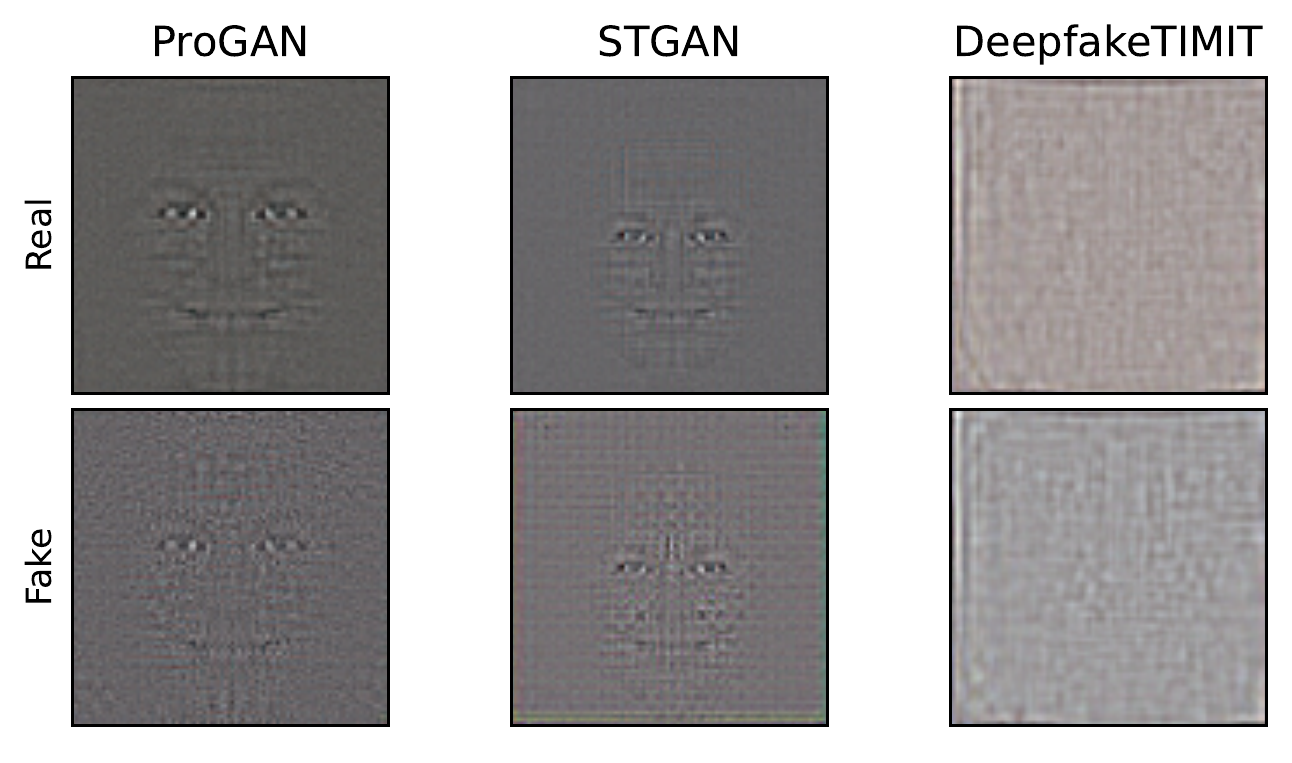}
    \caption{The empirical noise fingerprints of different DeepFake types estimated by average noise residual. The average noise residuals of the corresponding real images are also provided for comparison.}
    \label{fig:fingerprints}
\end{figure}

\begin{figure}[]
    \centering
    \includegraphics[width=0.47\textwidth]{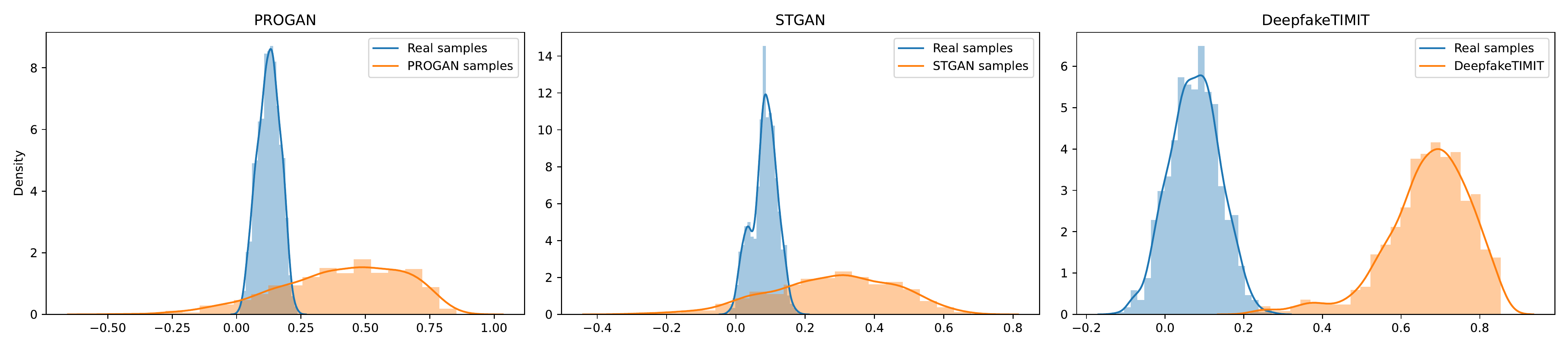}
    \caption{The distributions of correlations between individual samples and the noise fingerprints for different DeepFake types.}
    \label{fig:ncc}
\end{figure}

To show the fingerprints in the noise space, we estimate the fingerprints of different DeepFake types using the average noise residual. Specifically, the noise fingerprint of a DeepFake model $\mathcal{M}$ can be formulated as:

\begin{equation}
    F^{\mathcal{M}}=\frac{1}{N} \sum_{i=1}^{N} (I_{i}^{\mathcal{M}} - W(I_{i}^{\mathcal{M}})),
\end{equation}

where $I_{i}^{\mathcal{M}}$ is a sample generated by $\mathcal{M}$, $W(\cdot)$ is a Wiener denoising filter and $N=2000$ in our case. For comparison, we also used the same approach to calculate the average noise residual of an equal quantity of real images, as shown in Figure \ref{fig:fingerprints}. For all DeepFake types, the patterns of the average noise residuals between the real and fake samples are significantly different. Those of the real images have generally a smoother response than those of the fake images, which corroborates the fact that extra noise discrepancies are introduced into DeepFakes during their production. A distributional difference can also be demonstrated by calculating the normalized cross-correlation (NCC) between the average noise residual and the individual noise residuals from another $2000$ real/fake samples:
 
\begin{equation}
        \rho^{\mathcal{M}}_{i} = \frac{<F^{\mathcal{M}}, R_{i}^{\mathcal{M}}>}{\|F^{\mathcal{M}}\| \cdot \|R_{i}^{\mathcal{M}}\|},
\end{equation}
where $<\cdot, \cdot>$ and $\|\cdot\|$ denote the inner product and $l_2$-norm respectively; $R_{i} = I_{i}^{\mathcal{M}} - W(I_{i}^{\mathcal{M}})$. Figure \ref{fig:ncc} shows histograms of the individual correlations. For all DeepFake types, the NCC scores between the noise residuals of the real samples and the fingerprints of the DeepFakes are distributed around zero. This indicates that little correlation exists. By contrast, the NCC scores between the noise residuals of the fake samples and the fingerprints of the DeepFakes are remarkably larger than zero, testifying to a significant correlation with the corresponding fingerprint.

\subsection{Discussion}
The above model trace analysis identifies three typical model traces throughout the complete DeepFake pipeline. The interplay of these traces distinguishes DeepFake images from real ones. Hence, a solid and universal trace removal attack is desired to eliminate all these possible traces. In this way, the distribution of DeepFake images can be much closer to that of the real ones in the trace feature space, by which the modified DeepFake images can evade arbitrary detectors. Since our knowledge of DeepFake traces is derived from the fundamental pipeline shared by shared by the different methods used to generate DeepFakes, the trace removal attack  to be presented next is applicable to all these DeepFake types. 

\section{TR-Net: trace removal attack}

\subsection{Threat model}

\subsubsection{Victim model}
Assume the target victim model is an arbitrary DeepFake detector $\mathcal{C}$, which is a machine learning classifier that distinguishes trace features between real and DeepFake images. $\mathcal{C}$ takes an image $I$ or its hand-crafted features as input and outputs a binary decision of $\{Real, Fake\}$.

\subsubsection{Victim detector’s capability}
\label{sec:defense}
The attack is designed to defeat an arbitrary DeepFake detector $\mathcal{C}$. Therefore, there are few restrictions on $\mathcal{C}$’s capabilities. The developer of $\mathcal{C}$ can use discretionary model designs and feature engineering, and also sufficient training data. $\mathcal{C}$ is allowed to be trained with fake images from multiple DeepFake generation methods rather than a single one, so as to achieve better cross-task generalization and robustness.

In addition, more robust detectors are considered that have been embedded with defenses. As suggested in \cite{yu2019attributing, frank2020leveraging}, training data augmentation with perturbed images can significantly improve a detector’s robustness against common attacks. Hence, $\mathcal{C}$ is strengthened during its training phase with two augmentation strategies that confer two different levels of defense.

\textbf{Weak defense:} Empirical augmentation. This method adds perturbed samples from four empirical perturbation models following the settings in \cite{frank2020leveraging}:
\begin{itemize}
    \item Blurring: images are blurred with a Gaussian filter with a kernel size randomly sampled from $(3,5,7,9)$.
    \item Cropping: images are cropped along both sides with a random percentage sampled from $U(5,20)$ and then resized back to the original resolution.
    \item Compression: images are compressed with JPEG protocol with a quality factor randomly sampled from $U(10,75)$.
    \item Noising: i.i.d Gaussian noise is introduced into the images with a Gaussian variance randomly sampled from $U(5.0,20.0)$.
\end{itemize}
$\mathcal{C}$’s training set was augmented with a combination of these different perturbations in the order of: blurring, cropping, compression, noise. Each strategy was applied with a probability of $50\%$.

\textbf{Strong defense:} Adversarial augmentation. This method assumed that the developer of $\mathcal{C}$ had full knowledge of the attack model and could use the attack samples directly to augment the training set. This strategy was applied with a probability of $50\%$ as well.

\subsubsection{Attacker’s background knowledge}
The proposed attack requires little knowledge of $\mathcal{C}$, i.e., the attacker does not need to know the model architecture, parameters, or the features of $\mathcal{C}$’s interest. As such, there is no need to access the detector, its training set, or the query outputs.

To train the attack model, the attacker is assumed to have an auxiliary dataset containing real and fake images. Although this is a mild assumption given that there are plenty of ready-to-use DeepFake image datasets and models freely available to the public, some additional restrictions are still imposed on the attacker’s data availability to simulate the worst-case scenarios: 
\begin{itemize}
    \item Limited dataset size. The attacker has limited resources with which to collect public data and, thus, the resulting dataset size is relatively small.
    
    \item Out-of-distribution DeepFake. The attacker can only collect fake images generated by some particular DeepFake methods, which means the auxiliary dataset will not include all types of DeepFakes.
\end{itemize}

\subsubsection{Attack goals}
A successful attack means that the target detector $\mathcal{C}$ will be misled into classifying the attack samples as ‘$Real$’. Meanwhile, the attacker may expect the attack to be stealthy with preserving the visual utility of the original DeepFake image. Therefore, the visual difference between the original DeepFake image and the attack sample is required to be small enough that it would not be perceived by humans.

Formally, let $\mathbb{I}^{+}$ and $\mathbb{I}^{-}$ be the sets of real images and DeepFake images, respectively. Given a DeepFake image $I^{-} \in \mathbb{I}^{-}$, the attack model learns a mapping $\mathcal{A}: I^{-} \mapsto I^{*}$. The attacking sample $I^{*}$ satisfied the following attack goals:

(1) \textbf{Fraudulence.} The attack sample successfully deceives an arbitrary detector: $\forall{\mathcal{C}},  \quad  p(\mathcal{C}(I^{*}) = \mathcal{C}(I^{+})) \approx 1$;

(2) \textbf{Stealthiness.} The attack sample is perceptually indistinguishable from the original DeepFake image: $\forall{I^{-}}, \quad d(I^{-},I^{*})\leq \epsilon$, where $d(\cdot,\cdot)$ is a distance function.

\subsection{TR-Net}
\begin{figure}[]
    \centering
    \includegraphics[width=0.4\textwidth]{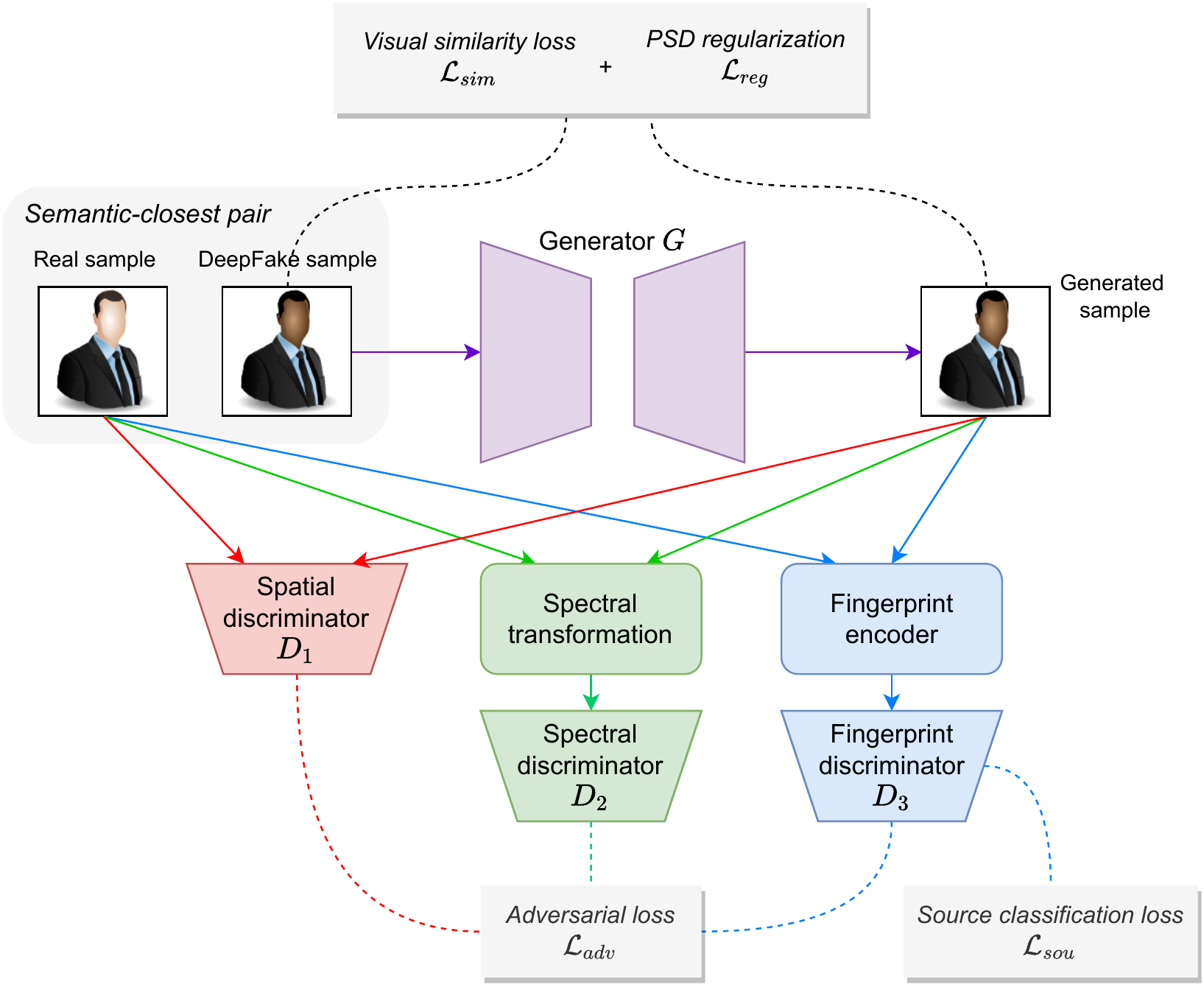}
    \caption{Framework overview of the trace removal network.}
    \label{fig:framework}
\end{figure}

Our trace removal attack is implemented with a trace removal network (TR-Net) based on adversarial learning. As shown in Figure \ref{fig:framework}, TR-Net consists of a generator $G$ and a set of discriminators $\mathbb{D}: \{D_{1}, D_{2}, D_{3}\}$. $G$ takes the original DeepFake images as inputs and reconstructs them to evade trace recognition by the discriminators. Each discriminator in $\mathbb{D}$ is devised for a specific auxiliary trace recognition task. Joint training on $\mathbb{D}$ adversarially impels $G$ to remove different traces concurrently. After the adversarial learning reaches Nash equilibrium, the optimal generator $G^{\star}$ is adopted as the attack model $\mathcal{A}$, i.e., $\mathcal{A} = G^{\star}$. Then, given a test DeepFake image $I^{-}_o$, the corresponding attack sample is $I^{\star}_o = A(I^{-}_o)$.

\subsubsection{Generator}
\begin{figure}[]
    \centering
    \includegraphics[width=0.45\textwidth]{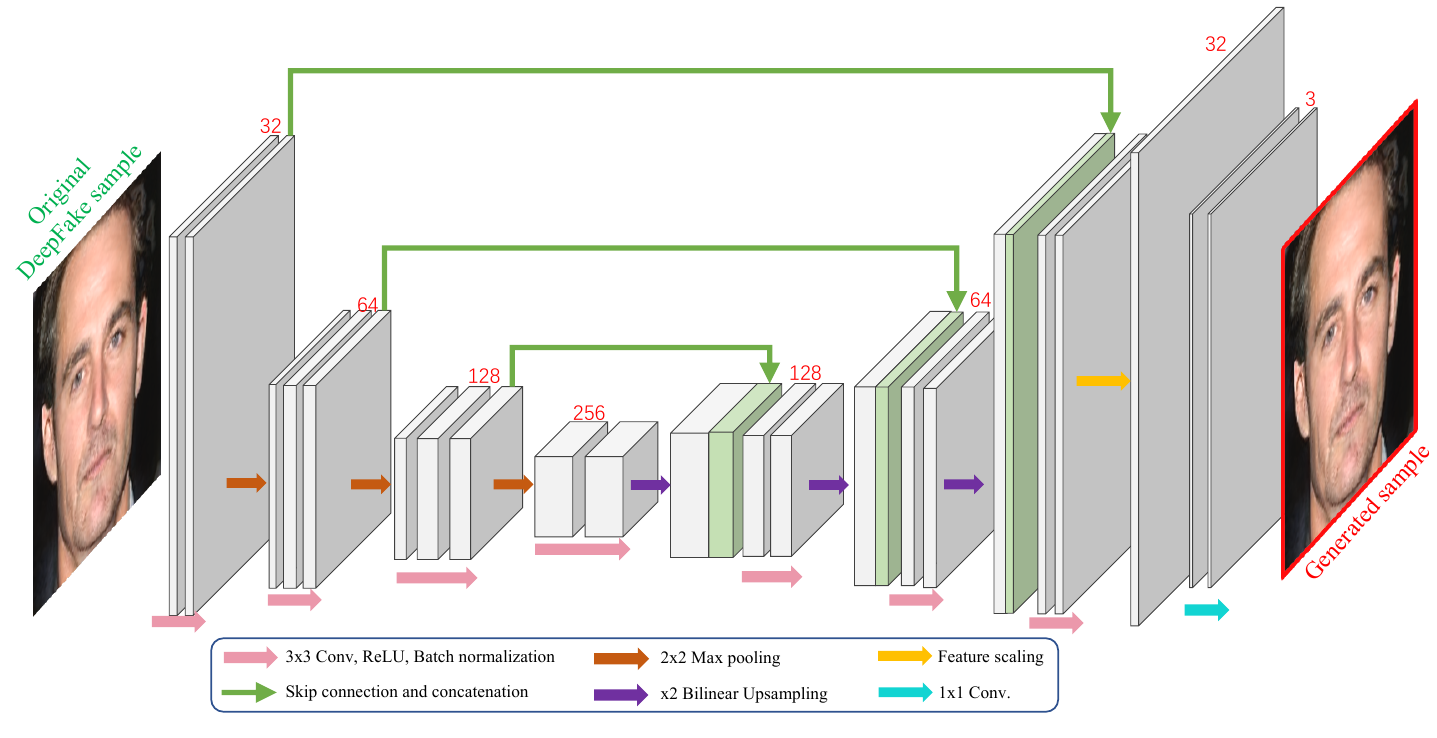}
    \caption{The network structure of the generator $G$}
    \label{fig:generator}
\end{figure}

The generator $G$ is a deep auto-encoder that learns to generate trace-free samples from the original DeepFake samples with an unchanged image size. The backbone of $G$ is a u-shaped network (U-Net) \cite{ronneberger2015u} given its remarkable capacity to reconstruct high-quality images. As shown in Figure \ref{fig:generator}, $G$ consists of an encoder path and a decoder path. The encoder involves repeated convolutional layers (with $3*3$ kernels) and max pooling layers (with $2*2$ kernels), capturing features at different scales of the images while compacting the spatial information. The decoder path is a symmetric expanding counterpart. In each decoding block, the feature map is upsampled to double size while the number of features is halved. Each decoder block also concatenates the output features with the high-resolution features from the corresponding encoder block, such that the feature and spatial information can be preserved for efficient reconstruction.
 
An additional challenge is that, as discussed in Section \ref{sec:DTD}, $G$ is a CNN-based generative model. Thus, it might produce its own model traces, which may interfere with the trace removal process. The loss functions proposed in the subsequent sections effectively suppress this intrinsic noise brought about by $G$. In addition, we also made two structural improvements as suggested in \cite{durall2020watch} and \cite{chandrasegaran2021closer} to mitigate this problem. First, we replaced the transposed convolution-based upsampling in the original U-Net with bi-linear interpolation-based upsampling. Second, we added a feature scaling layer before the last convolutional layer of $G$.

\begin{figure}[]
    \centering
    \includegraphics[width=0.45\textwidth]{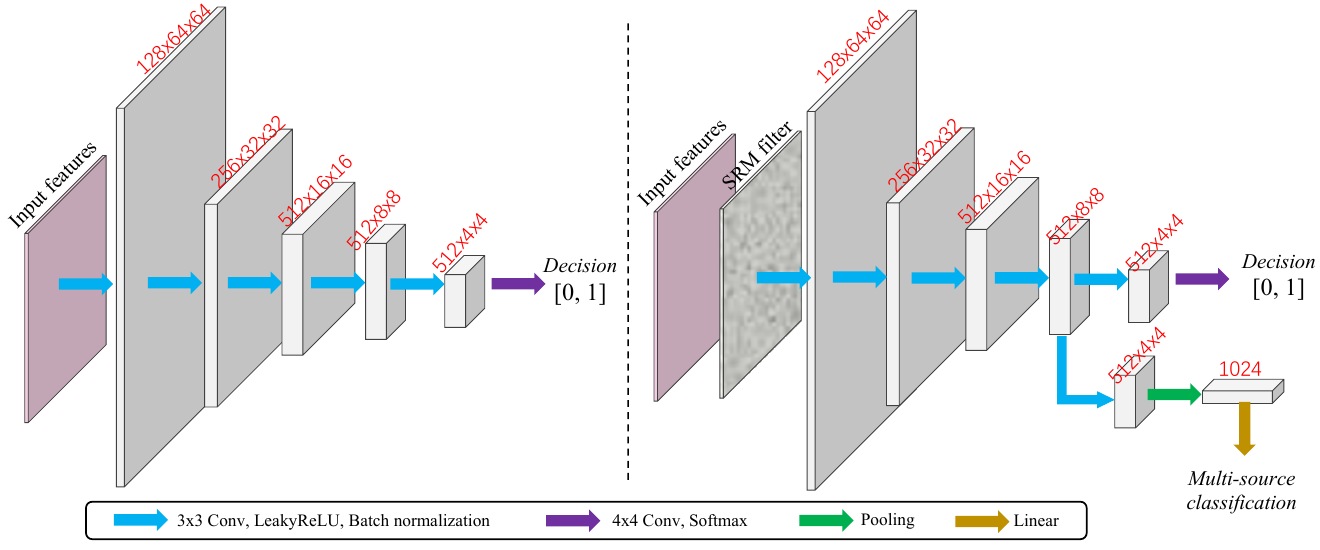}
    \caption{The network structures of the discriminators. The left one is used for the spatial discriminator $D_{1}$ and the spectral discriminator $D_{2}$, and the right one with an additional SRM filter layer and a multi-task head layer is used for the fingerprint discriminator $D_{3}$.}
    \label{fig:discriminator}
\end{figure}

\subsubsection{Discriminators}
Discriminator $D$ impels $G$ to produce trace-free attack samples via adversarial learning. As a result, $D$ needs to be able to recognize accurate DeepFake trace patterns by learning to classify real and fake images in the trace space. 
According to our trace discovery, three types of traces are revealed in different domains, each with a unique representation. The interplay and inter-domain interference across traces make a single discriminator learned in a single feature subspace impractical to represent every trace accurately. To this end, we propose disentangling the trace representations in the input space and employing a set of parallel discriminators $\mathbb{D}: \{D_{1}, D_{2}, D_{3}\}$ to leach the different traces separately.
\noindent \paragraph{Spatial discriminator $D_{1}$} 
$D_{1}$ captures potential spatial anomalies in the spatial domain, including distortions, inconsistencies, disharmony, etc. Similar to the original discriminator in a normal GAN, $D_{1}$ is trained directly with the RGB pixel values, and thus can be seen as an incremental refinement on the raw DeepFake images in terms of visual quality. $D_{1}$ is built on a five-layer CNN, as shown in Figure \ref{fig:discriminator}. Note that using a complicated structure for the discriminator is unnecessary. In addition to extra computational cost, a complicated discriminator leads to an imbalance between the generator and discriminators during training. A shallow CNN is sufficient to capture these traces accurately in our experiments. 

\noindent \paragraph{Spectral discriminator $D_{2}$}
$D_{2}$ learns to recognize the spectral disparities between the real and attack samples. $D_{2}$ uses the same CNN backbone as $D_{1}$, but takes the frequency spectrum instead of RGB pixels as its input. The frequency spectrum is transformed from the pixel values by two-dimensional Discrete Fourier Transform (2D-DFT). Given a natural image $I \in \mathbb{R}^{M \times N}$, the 2D-DFT maps each pixel value of the gray-scale component of $I$ to a frequency value $\mathcal{F}(u, v) \in \mathbb{R}^{M\times N}$: 

\begin{equation}
\small
\begin{aligned}
\mathcal{F}(I)(u,v)=\sum_{m=0}^{M-1} \sum_{n=0}^{N-1} I(m, n) \cdot e^{-2 \pi i \cdot (\frac{u m}{M} + \frac{v n}{N})}. 
\end{aligned}
\end{equation}

As the imaginary part is incompatible with a CNN for calculating gradients, directly applying the 2D-DFT $\mathcal{F}(I)$ to $D_{2}$ is impractical. Instead, we decompose the complex-valued matrix of $\mathcal{F}(I)$ into its amplitude response $\mathcal{F}_{am}(I)$ and phase response $\mathcal{F}_{ph}(I)$. Let the complex form of $\mathcal{F}(I)$ be $\mathcal{F}(I) = a + bi$, and we have:

\begin{equation}
\small
\begin{aligned}
 \mathcal{F}_{am}(I) = |\mathcal{F}(u, v)|=\sqrt{a^{2}+b^{2}} \\
 \mathcal{F}_{ph}(I) = \angle \mathcal{F}(u, v)=\arctan \frac{b}{a}.
\end{aligned}
\end{equation}

Then, the two components are concatenated as a 2-channel real-valued matrix as the input of $D_{2}$, denoted as $\widehat{I} = [\mathcal{F}_{am}(I), \mathcal{F}_{ph}(I)]$.

\noindent \paragraph{Fingerprint discriminator $D_{3}$}
$D_{3}$ targets the DeepFake’s model fingerprint in the noise space. As the trace representations need to be disentangled in the input feature space, the fingerprint should be extracted before learning the binary classification and, for this, a reliable fingerprint encoder is required. Inspired by two existing DeepFake fingerprint encoders, one is a noise-based encoder \cite{marra2019gans} and the other is a learning-based encoder \cite{yu2019attributing}, we propose to combine the two insights together to obtain a better fingerprint encoder. Then, the encoding pipeline is incorporated with $D_{3}$ via multi-task learning for end-to-end training. 

First, a residual noise extraction is performed to represent the statistical fingerprints. We opted for an SRM filter for this purpose given its reliability in estimating local noise distributions for image forensics \cite{fridrich2012rich}. The input to $D_{3}$ is then denoted as $\widetilde{I} = \operatorname{SRM}(I)$. A three-layer SRM filter is concatenated to the bottom of $D_{3}$, with the following kernels:

\scriptsize
\begin{equation}
\setlength{\arraycolsep}{0.8pt}
\nonumber k_1=\frac{1}{4}{
\begin{bmatrix}
0 & 0 & 0 & 0 & 0\\
0 & -1 & 2 & -1 & 0\\
0 & 2 & 4 & 2 & 0\\
0 & -1 & 2 & -1 & 0\\
0 & 0 & 0 & 0 & 0\\
\end{bmatrix}
},
k_2=\frac{1}{12}{
\begin{bmatrix}
-1 & 2 & -2 & 2 & -1\\
2 & -6 & 8 & -6 & 2\\
-2 & 8 & -12 & 8 & 2\\
2 & -6 & 8 & -6 & 2\\
-1 & 2 & -2 & 2 & -1\\
\end{bmatrix}
}, \\
k_3=\frac{1}{2}{
\begin{bmatrix}
0 & 0 & 0 & 0 & 0\\
0 & 0 & 0 & 0 & 0\\
0 & 1 & -2 & 1 & 0\\
0 & 0 & 0 & 0 & 0\\
0 & 0 & 0 & 0 & 0\\
\end{bmatrix}
}
\end{equation}

\normalsize
Then, a two-layer fingerprint encoding head is connected to the penultimate layer of $D_{3}$ to perform multi-task learning. According to \cite{yu2019attributing}, the encoding head is trained in a DeepFake source identification task for unique fingerprint representations. Since the real images are also included as a possible source in this task, this multi-source classification can be seen as a fine-grained representation compared to the coarse “real-or-fake” classification. In this way, jointly learning the two head tasks gives synergistic effects for a better fingerprint representation.

\subsubsection{Loss functions}
\begin{figure}[]
    \centering
    \includegraphics[width=0.3\textwidth]{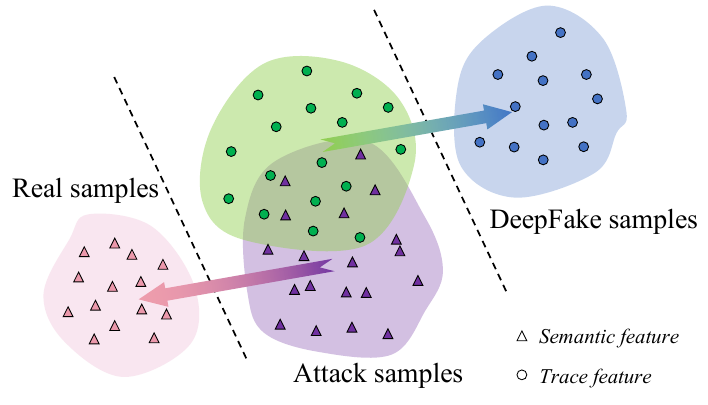}
    \caption{The diagram of the changes in the latent feature space of TR-Net during optimization. A conflict in feature migration directions occurs owing to the overlap of trace features and semantic features.}
    \label{fig:features}
\end{figure}

We design an adversarial loss to supervise both $G$ and $\mathbb{D}$ of the TR-Net, which can enable trace removal so as to realize the attacking goal of fraudulence. Regarding the goal of stealthiness, a visual similarity loss is imposed on $G$ to ensure that the semantic information of the original DeepFake samples are perfectly preserved in the corresponding attack samples. In addition to achieving these attack goals, one technical challenge is that an ideal trace removal attack requires simultaneously closing the distribution gap between: the attack samples and the real samples at the trace level; and between the attack samples and the DeepFake samples at the semantic level (see Figure \ref{fig:features}). However, due to the information continuity in an image, the trace features inevitably overlap the semantic features in the latent space, leading to a potential conflict in feature migration directions during optimization. Our loss function design mitigates this nontrivial problem, as shown next.

\noindent \paragraph{Adversarial loss} 
The adversarial learning of TR-Net is performed with the input data pairs in the form of $(I^{+}, I^{-})$. The discriminators continuously learn to distinguish the generator’s output $G(I^{-})$ from $I^{+}$ in different feature spaces, while the generator tries to mislead the discriminators’ judgements about $G(I^{-})$. Conventionally, $I^{+}$ and $I^{-}$ are randomly sampled from $\mathbb{I}^{+}$ and $\mathbb{I}^{-}$ respectively. However, the random sampling is less practical for TR-Net’s optimization considering the conflicts between semantic features and trace features. The visual information between $I^{+}$ and $I^{-}$ should be as consistent as possible to enforce the discriminators to focus on purer trace features while reducing their bias to semantic features. Thus, the \textit{semantically-closest pairs} are constructed to supervise discriminators. If a fake sample is produced by a method where a real source image exists, such as facial attribute editing or face replacement, the source image is applied straightforwardly as the semantically-closest counterpart. For a face synthesis sample created out of nowhere, its nearest neighbor is retrieved from the real image set $\mathbb{I}^{+}$ as a counterpart.

With the semantically-closest pair $(I^{+}, I^{-})$ in hand, the adversarial loss for jointly training the “one-versus-multiple” framework is denoted as:

\begin{equation}
\small
    \mathcal{L}_{adv}(G, D_{1}, D_{2}, D_{3}) = \mathcal{L}(G, D_{1}) + \mathcal{L}(G, D_{2}) +  \mathcal{L}(G, D_{3}),
\end{equation}
where
\begin{equation}
\small
\mathcal{L}(G, D_{i}) = \mathbb{E}_{x^{+}, x^{-}}[log(D_{i}(x^{+})) + log(1 - D_{i}(G(x^{-})))],\\
\end{equation}
The input $x$ varies for different discriminator, i.e., $x=I$ for $D_{1}$, $x=\widehat{I}$ for $D_{2}$ and $x=\widetilde{I}$ for $D_{3}$. Since different traces are deemed as equally significant, no additional weights are applied to each sub-item of $\mathcal{L}_{adv}$.   

\noindent \paragraph{Visual similarity loss} 
To satisfy the stealthiness goal, a visual similarity loss is additionally imposed on the generator $G$. The commonly-used pixel-wise distance $||I^{-} - G(I^{-})||_{2}$ is not particularly applicable to our method as it will typically lead to overfitting the visual information. In turn, this will exacerbate the conflict between the semantic features and the trace features, thus compromising the trace removal. Moreover, despite having $D_{2}$ to encourage spectra matching from the attack samples to the real images, we experimentally find that only a $D_{2}$ is insufficient to well match high-frequency components. This is because in natural images, information tends to be centralized in lower-frequency components.

Instead, we propose a novel visual similarity loss plus a power spectral density (PSD) regularization to cope with the above problem. Given an image $I$, first, a filter is applied to its center-shifted DFT spectrum. This decomposes $I$ into its low frequency components $I_{l}$ and high frequency components $I_{h}$:
\begin{equation}
\small
\left\{ 
    \begin{array}{l}
    I_{l} = \mathcal{F}^{-1}(\mathcal{H}(u,v)\cdot\mathcal{F}(u,v)) \\
    I_{h} = \mathcal{F}^{-1}(1-\mathcal{H}(u,v)\cdot\mathcal{F}(u,v))
    \end{array}
\right.,
\end{equation}
where $\mathcal{F}^{-1}$ is the reverse DFT, $\mathcal{H}(u, v)= exp(-\frac{u^{2}+v^{2}}{2 \sigma^{2}})$ is a Gaussian filter. Then the visual similarity loss between a source fake image $I^{-}$ and its reconstructed version $G(I^{-})$ is computed as the VGG perceptual loss \cite{johnson2016perceptual} on the low frequency components:
    


\begin{equation}
\small
    \mathcal{L}_{sim}(G)= \frac{1}{W*H}\left\|\operatorname{VGG}_{k}\left(I^{-}_{l}\right)-\operatorname{VGG}_{k}\left(G(I^{-})_{l}\right)\right\|^{2}_{2},
\end{equation}
where $W$ and $H$ are the dimensions of the respective feature maps within the VGG network \cite{simonyan2014very} and $\operatorname{VGG}_{k}$ denotes the features extracted at VGG’s $k$-th layer. 

Additionally, a PSD regularization is added to the visual similarity loss to enforce the mapping of frequency information between the attack samples and the real images. The PSD of an image $I$ can be represented as a one-dimensional profile of the center-shifted power spectrum resulting from an azimuthal integration over each radial frequency $\theta$:

\begin{equation}
\small
\begin{aligned}
&\operatorname{PSD}\left(\omega_{k}\right)=\int_{0}^{2 \pi}\left\|\mathcal{F}(I)\left(\omega_{k} \cdot \cos (\theta), \omega_{k} \cdot \sin (\theta)\right)\right\|^{2} \mathrm{~d} \theta \\
&\text { for } \quad k=0, \ldots, M / 2-1.
\end{aligned}
\end{equation}
Benefiting from the semantically-closest pair $(I^{+},I^{-})$ where the lower frequency components are close to each other, the PSD regularization can operate on the high frequency components merely, which is computed as the Euclidean distance between the PSDs of $I^{+}_{h}$ and $G(I^{-})_{h}$: 
\begin{equation}
\small
\mathcal{L}_{reg}(G)= \frac{1}{M/2-1}\left\|\operatorname{PSD}(I^{+}_{h}) - \operatorname{PSD}(G(I^{-})_{h}) \right\|_{2}^{2} 
\end{equation}

\noindent \paragraph{Source identification loss} A source identification loss is required specifically to train the auxiliary source identification head of $D_3$. The multi-classification cross entropy loss is applied here:
\begin{equation}
\small
    \mathcal{L}_{sou}(D_3) =-\sum_{s}^{S} y^{s} \log D_3(I^{(s)}), I^{(s)} \in \{\mathbb{I^{+}}, \mathbb{I^{-}}\}
\end{equation}
where $S$ is the total count of sources in the training dataset, $y^{(s)}$ and $D_3(I^{(s)})$ are the ground-truth and $D_3$-predicted source labels for the sample $I^{(s)}$, respectively. 

The final training objective of the TR-Net is:
\begin{equation}
\small
\begin{aligned}
    T = arg \min _{\{G, D_{3}\}} \max _{\{D_{1}, D_{2}, D_{3}\}}\left\{\mathcal{L}_{adv} + \mathcal{L}_{sim} + \lambda \mathcal{L}_{reg} +  \mathcal{L}_{sou} \right\}
\end{aligned}
\end{equation}

\subsection{Comparison with previous attacks}
To date, the published adversarial attacks have had some limitations. First, searching for the optimal adversarial noise perturbations to a target detector typically requires a certain level of information about the detector itself, such as the parameters, network structure, or the outputs. Thus, there will be a transferability issue when facing an unknown or black-box detector \cite{barni2019transferability, zhao2020effect}. Second, the feasibility of an adversarial attack on some advanced detectors which involve sophisticated network designs will be problematic. Under these circumstances, it becomes difficult, if not impossible, to search for the optimal perturbations that will maintain a high attack success rate while being largely imperceptible.

Regarding the reconstruction-based attacks, our method is analogous to this genre of attacks, but fundamental differences exist. Similar to adversarial attacks, reconstruction-based attacks are performed in a “detector-specific” way. The attacker is assumed to know what type of forgery features are of prime interest to the target detector. What is worse, these attacks solely focused on an individual feature type in a single signal domain, irrespective of the fact that various traces exist in different domains.

By contrast, our method improves on anti-forensic attacks by removing multiple forgery traces at the same time. Additionally, they are removed in a way that is agnostic to the detector. The result is better transferablity to an unknown detector. Technically, this is more challenging than dealing with a single trace feature given the interplay between traces and the inter-domain interference, yet the proposed TR-Net is competent to meet the challenge.

\section{Experimental Evaluations}
\label{sec:experiment}
We evaluated the proposed trace removal attack in heterogeneous security scenarios where the attacker has different background knowledge of the data, and the detectors’ defensive capability varies. In each scenario, the effectiveness of the attack was assessed by verifying whether the two goals, fraudulence and stealthiness, had been satisfied when attacking six representative detectors. Alongside validating the goals, we additionally provide a closer look into the trace removal result from different dimensions to justify its success. 

\subsection{Evaluation metrics}

(1) The fraudulence goal is verified in terms of detection accuracy, calculated as the proportion of correctly classified samples out of all the samples in a single class. Attack samples with higher fraudulence result in lower detection accuracy of the test detector.

\noindent (2) The stealthiness goal is verified by assessing the visual quality loss in attack samples. We used peak signal-to-noise ratio (PSNR) and structural similarity (SSIM) between an original DeepFake image and its corresponding attack sample to evaluate the visual quality loss. PSNR quantifies the ratio between the maximum possible power of a signal and the power of corrupting noise that affects the fidelity of its representation. SSIM is a common metric for measuring the similarity between two images. A larger value in either PSNR or SSIM indicates a smaller loss in visual quality, which equates to better stealthiness of the attack sample.

\subsection{Datasets}
The proposed trace removal attack is applicable to all DeepFake types described in Section \ref{sec:deepfake}, including face synthesis, facial attribute editing, face replacement. To the best of our knowledge, the existing publicly-available DeepFake detection datasets, fail to cover all these methods. Thus, for a thorough evaluation, we created \textit{All-in-One-DF}, a new DeepFake dataset based on previous datasets.

The \emph{All-in-One-DF} dataset consists of $66,000$ semantically-closest pairs of real and fake images (i.e., $132,000$ images in total) from four sources:

(1) CelebA: A large-scale dataset containing more than $200k$ real face images. The images are cropped and aligned to the size of $128*128*3$ with the face in the centre.

(2) Face synthesis: We employed ProGAN, one of the most popular unconditional GANs to synthesize non-existing face images. We utilize the pre-trained\footnote{\label{footnote_pretrain}Both the ProGAN and STGAN instances are pre-trained on the CelebA dataset and thus there is no domain discrepancy between the original ProGAN and STGAN images.} ProGAN instance \cite{yu2019attributing} to generate $22,000$ fake images. Then we retrieval their corresponding $1$-nearest-neighbor similar counterparts from the CelebA dataset to construct semantically-closest pairs. 

(3) Facial attribute editing: We selected STGAN, a state-of-the-art GAN for facial attribute editing for this use. We randomly sampled $22,000$ real images from the remaining CelebA dataset and applied the official pre-trained\textsuperscript{\ref{footnote_pretrain}} STGAN instance \cite{liu2019stgan} to modify the attributes on these real samples, resulting in $22,000$ fake samples. To ensure the attribute diversity, we modified the soft-biometric attribute (facial age) for half of the images and the appearance attribute (hair colour) for the remaining half. 

(4) Face replacement: DeepfakeTIMIT \cite{korshunov2018deepfakes} is a human video dataset where faces are swapped and rendered using GAN-based approaches. There are $320$ pairs of source videos and their face-swapped counterparts in DeepfakeTIMIT. We randomly selected $22,000$ frames from all videos on either side, followed by face-centered cropping to the size of $128*128*3$.  

\begin{figure}[]
    \centering
    \includegraphics[width=0.3\textwidth]{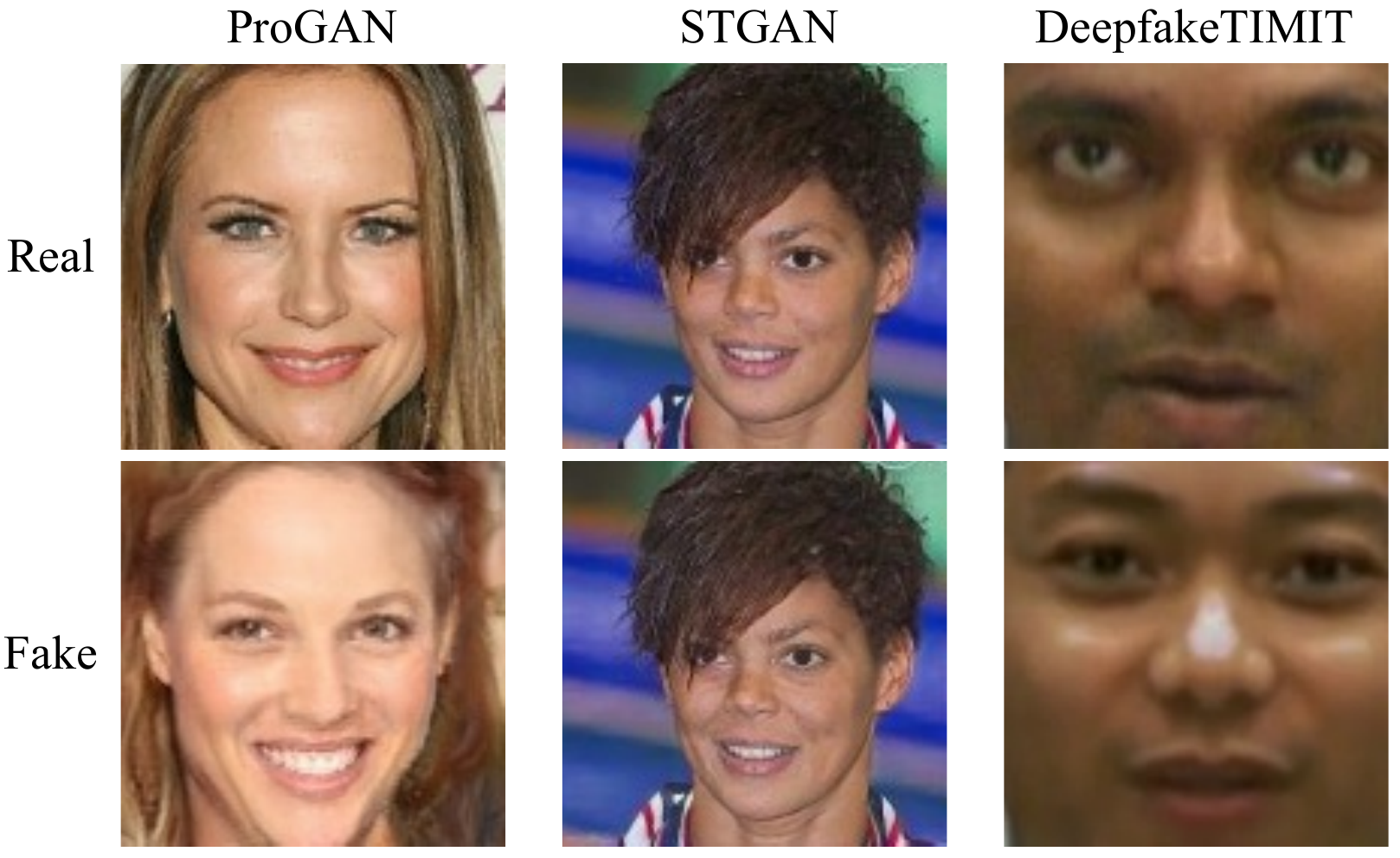}
    \caption{Examples of semantically-closest pairs from different DeepFake sources.}
    \label{fig_pairs}
\end{figure}

Figure \ref{fig_pairs} provides some pairwise examples from the \emph{All-in-One-DF} dataset. 

\subsection{Selected Victim Detectors}
To show the efficacy of the proposed attack model in attacking arbitrary detectors, we selected six representative DeepFake detectors recently proposed in the literature, which evenly cover the three detector categories outlined in Section \ref{sec:detectors}.

Spatial-based detectors:
\textbf{Xception \cite{rossler2019faceforensics++}} is a deep CNN widely adopted as the backbone network in face forgery forensics tasks. It has achieved leading performance in some benchmark datasets by learning directly from RGB pixel inputs. \textbf{Patch-CNN \cite{chai2020makes}} focuses on the local properties in semantic regions rather than on global semantics. It aggregates the decisions of a set of truncated Xceptions learned from image patches for the final binary decision. 

Frequency-based detectors: \textbf{DCTA \cite{frank2020leveraging}} is a shallow CNN classifier learned from the 2D-DCT spectra of images. \textbf{F$^{3}$-Net \cite{qian2020thinking}} is one of the state-of-the-arts in DeepFake detection. It involves a two-stream collaborative network that combines frequency-aware decomposition and local frequency statistics to learn frequency-aware clues.

Fingerprint-based detectors: \textbf{LF \cite{yu2019attributing}} is a deep CNN that learns GAN fingerprints in a multi-source identification task. The original multi-classification results are further divided into the “real-or-fake” binary decisions. \textbf{NF \cite{marra2019gans}} is a non-trainable method that differentiates GAN images from real ones via a cross correlation score of the noise residual-based fingerprints.

\subsection{Settings}
The \emph{All-in-One-DF} dataset is randomly partitioned into a training set with $60,000$ semantically-closest pairs and an evaluation set with $6,000$ pairs. For all detectors, we followed the training settings recommended in the original papers. The detectors were trained on the training set with a $9:1$ training-validation ratio. Regarding the training of TR-Net, we set the batch size to $150$. Both the generator and discriminators were optimized using the RMSprop optimizer \cite{ruder2016overview} with initial learning rates of $1.6e-3$ and $1.6e-4$, respectively, plus a scheduler with a decay rate of $0.5$. The scheduler was executed at the end of a training epoch if the loss stopped decreasing. There were $8$ training epochs in total. The weight $\lambda$ was set to $100$. After training, the checkpoint with the minimal generator loss in the last epoch was nominated as the attack model and was used on the $6,000$ fake images in the evaluation set to craft attack samples.

\subsection{Attacking with unlimited background knowledge}
We first evaluate the attack performance in the scenario where the attacker has no limits on the background knowledge of data, i.e., the whole training set is available for training the attack model. For a comprehensive evaluation, the proposed trace removal attack is compared with several baseline attack methods. Also the detectors with varying abilities were considered, i.e., detectors without defense, with weak defense, and with strong defense.  
    
\subsubsection{Baseline attacking methods}
We included four other attack methods to demonstrate baseline performance, including adding random noise (\textbf{Noise}), two classic adversarial attacks \textbf{FGSM} \cite{kurakin2018adversarial} and \textbf{PGD} \cite{madry2017towards}, and a reconstruction-based attack \textbf{GANprintR} \cite{neves2020ganprintr}. The \textbf{Noise} operation is the same as the "noising" perturbation described in Section \ref{sec:defense}. Both the \textbf{FGSM} and \textbf{PGD} attacks were optimized based on the \textbf{Xception} detector and then applied to all detectors so as to assess the white-box and black-box attack capacities simultaneously. The maximum perturbation $\epsilon$ was set as $0.003$ for both attacks. For \textbf{GANprintR}, we followed the setting in the original paper.

\subsubsection{Evaluating fraudulence}
\paragraph{Attacking detectors without defenses}
Table \ref{tab_wo_def} details the attack results against detectors without defense in terms of classification accuracy. The first two columns show the real and fake classification accuracy before being attacked. All detectors here reached a satisfactory level of accuracy at over $90.00\%$, except for the \textbf{NF} detector. This is because \textbf{NF} is a rule-based detector while the others are learnable ones. After attack, the accuracy of all detectors decreased. This shows that all detectors, even the current state-of-the-arts, are vulnerable to attacks, and perfect detection robustness is still a long-term challenge. Comparatively, the frequency-based detectors were more resistant to the attacks than the others, especially \textbf{F$^{3}$-Net}, which benefits from a sophisticated frequency-aware feature engineering. We suppose the superiority of frequency-based detectors is attributed to the nature that all attack models, whether noise-adding or reconstruction-based ones, will inevitably lead to detectable changes in the frequency domain.

Among the baseline attacks, the two adversarial attack methods, \textbf{FGSM} and \textbf{PGD}, were more effective than the others, and particularly destructive to \textbf{Xception}. This is not surprising because these two methods are optimized specifically for \textbf{Xception} in a white-box manner. They also showed a certain level of transferability when facing \textbf{Patch-CNN}, given that \textbf{Patch-CNN} and \textbf{Xception} both rely on pixel inputs and have structural similarity in terms of Xception blocks. However, as earlier discussed, the detector-specific design leads to poor transferability of \textbf{FGSM} and \textbf{PGD} on other types of detectors unknown to the two attack models. By comparison, \textbf{TR-Net} took advantage of the detector-agnostic design, achieving competitive or superior results in attacking all six detectors. After the trace removal attack, the classification accuracy of all detectors had decreased markedly, and the average accuracy of the six had dropped from $90.89\%$ to $25.66\%$. The results indicate the proposed trace removal attack is universal and well transferable across different detectors.

\begin{table}[]
  \centering
  \scriptsize
  \caption{Attack performances of five attack methods against six detectors without defenses. The lowest detection accuracy in each line is in bold.}
  \renewcommand\arraystretch{1.2}
  \setlength{\tabcolsep}{1mm}{
  
    \begin{tabular}{l|ccccccc}
    \toprule
         Accuracy(\%) & \textbf{Real} & \textbf{Fake} & \textbf{Noise} & \textbf{FGSM} & \textbf{PGD} & \textbf{GANprintR} & \textbf{TR-Net} (ours) \\
    \midrule
    \textbf{Xception} & 99.95  & 99.86  & 65.44  & 4.43  & \textbf{0.01} & 58.53  & 18.90  \\
    \textbf{Patch-CNN} & 93.47  & 92.13  & 53.91  & 12.36  & \textbf{9.81} & 57.31  & 13.06  \\
    \textbf{DCTA} & 95.51  & 90.66  & 51.59  & 33.18  & 25.37  & 70.24  & \textbf{30.16} \\
    \textbf{F$^3$-Net} & 100.00  & 99.97  & 85.41  & 49.62  & \textbf{45.73} & 80.73  & 56.10  \\
    \textbf{LF} & 94.66  & 91.55  & 37.12  & 16.00  & 15.55  & 64.76  & \textbf{14.01} \\
    \textbf{NF} & 78.81  & 71.12  & 42.65  & 28.21  & 25.70  & 31.88  & \textbf{21.74} \\
    \textbf{Average} & 93.74  & 90.89  & 56.02  & 23.97  & \textbf{20.36} & 60.58  & 25.66  \\
    \bottomrule
    \end{tabular}}%
  \label{tab_wo_def}%
\end{table}%

\paragraph{Attacking detectors with defenses}
Next, we tested the attacks in the cases that the detectors are embedded with varying degrees of defense as described in Section \ref{sec:defense}. Detectors with the weak defense, i.e., the empirical augmentation strategy, are denoted as \{\textbf{model name}\}\textbf{(+)} and those with the strong adversarial augmentation defense are denoted as \{\textbf{model name}\}\textbf{(++)}.

\textbf{Weak defense}. Table \ref{tab_weak_def} shows the classification accuracy of detectors embedded with the empirical data augmentation strategy. The high accuracy values in the first two columns indicate that all detectors maintained stable detection on clean samples, despite some slight but acceptable decreases e.g., \textbf{Xception(+)} or \textbf{F$^3$-Net(+)}. This is a side effect of perturbation-based data augmentation.

We can see that after being strengthened with empirical data augmentation, the robustness of all detectors were improved against all attack methods. Regarding the attack methods, the \textbf{Noise} attack is barely misled the detectors, and the attacking specificity of \textbf{FGSM} and \textbf{PGD} on \textbf{Xception(+)} and \textbf{Patch-CNN(+)} was no longer significant. Comparatively, the \textbf{TR-Net} maintained satisfactory, degrading the classification accuracy of almost all detectors to lower than random guess, except for \textbf{F$^3$-Net(+)}. Moreover, \textbf{TR-Net} surpassed all baseline attack methods in five out of the six detector groups.

\begin{table}[]
  \centering
  \scriptsize
  \caption{Attack performances of five attack methods against six detectors with weak defenses. The lowest detection accuracy in each line is in bold. Red color indicates the white-box attack result.}
  \renewcommand\arraystretch{1.2}
  \setlength{\tabcolsep}{1mm}
    \begin{tabular}{l|ccccccc}
    \toprule
        Accuracy(\%)  & \textbf{Real} & \textbf{Fake} & \textbf{Noise} & \textbf{FGSM} & \textbf{PGD} & \textbf{GANprintR} & \textbf{TR-Net} (ours) \\
    \midrule
    \textbf{Xception +} & 98.55  & 98.86  & 98.00  & {\color{red}34.73}  & {\color{red}33.35}  & 73.37  & \textbf{31.13} \\
    \textbf{Patch-CNN +} & 96.27  & 90.66  & 90.01  & 58.47  & 42.19  & 79.65  & \textbf{33.84} \\
    \textbf{DCTA +} & 98.40  & 94.99  & 95.77  & 46.94  & 45.44  & 84.80  & \textbf{44.66} \\
    \textbf{F$^3$-Net +} & 99.92  & 99.64  & 96.87  & \textbf{55.79} & 56.16  & 91.75  & 73.03  \\
    \textbf{LF +} & 94.57  & 94.63  & 82.70  & 48.68  & 39.91  & 76.54  & \textbf{35.60} \\
    \textbf{NF +} & 80.43  & 75.54  & 74.99  & 59.71  & 56.62  & 58.70  & \textbf{40.84} \\
    \textbf{Average} & 94.69  & 92.39  & 89.72  & 50.72  & 45.61  & 77.47  & \textbf{43.18} \\
    \bottomrule
    \end{tabular}%
  \label{tab_weak_def}%
\end{table}%

\textbf{Strong defense.} Table \ref{tab_strong_def} shows the classification accuracy of the detectors embedded with the adversarial data augmentation strategy. Note that unlike the weak defense where all detectors share the same empirical augmentation strategy, in the strong defense, the adversarial augmentation strategies are specific to the attack methods, and thus the results on clean (\textit{cle}) / attacked (\textit{att}) DeepFake samples are reported individually for each attack method. 

From the table, we can see that the adversarial data augmentation strategy substantially improved the robustness of all detectors against the four baseline attack methods. Take \textbf{PGD}, the best baseline attack method in our experiments as an example, the average accuracy of strongly defended detectors only degraded from $91.22\%$ to $83.28\%$, whereas the corresponding result for the weakly-defended detector is $92.39\%$ down to $55.28\%$, and $90.89\%$ down to $27.48\%$ for the naked detector. However, the strong defense only had a relatively small impact on our attack method. The average detection accuracy on the \textbf{TR-Net} attack samples was much lower than that of other types of attack samples, at merely a little higher than a random guess.

We also observe an intriguing phenomenon that, as shown in Table \ref{tab_strong_def}, adversarial augmentation with the \textbf{TR-Net} samples dramatically compromised the detectors’ ability to classify clean fake samples, which is inconsistent with the impacts of the augmentations with other types of attack samples or the empirical augmentation. The reason may be that the samples after trace removal are inherently closer to the real samples, which may have confused the detector during training. This reveals the potential to use trace removal attack to poison a DeepFake detection dataset, which remains future investigation.


\begin{table}[]
  \centering
  \tiny
  \caption{Attack performances of five attack methods against six detectors with strong defenses. The detection accuracy on clean (cle) / attacked (att) samples are shown individually for each attack method. The lowest accuracy in each line is in bold. Red color indicates the white-box attack result.}
  \renewcommand\arraystretch{1.2}
  \setlength{\tabcolsep}{1.5mm}
    \begin{tabular}{l|cccccccccc}
    \toprule
    \multicolumn{1}{c}{}  & \multicolumn{2}{c}{\textbf{Noise}} & \multicolumn{2}{c}{\textbf{FGSM}} & \multicolumn{2}{c}{\textbf{PGD}} & \multicolumn{2}{c}{\textbf{GANprintR}} & \multicolumn{2}{c}{\textbf{TR-Net} (ours)} \\
\cline{2-11}    \multicolumn{1}{c}{}  & \textit{cle} & \textit{att} & \textit{cle} & \textit{att} & \textit{cle} & \textit{att} & \textit{cle} & \textit{att} & \textit{cle} & \textit{att} \\
    \hline
    \textbf{Xception ++} & 99.37  & 99.64  & 97.69  & {\color{red}71.22}  & 97.36  & {\color{red}59.32}  & 98.34  & 98.37  & 60.17  & \textbf{49.67} \\
    \textbf{Patch-CNN ++} & 94.83  & 94.36  & 93.51  & 82.29  & 92.96  & 76.26  & 96.25  & 95.86  & 55.53  & \textbf{41.37} \\
    \textbf{DCTA ++} & 93.08  & 91.20  & 93.53  & 77.87  & 92.88  & 78.91  & 94.65  & 96.32  & 77.45  & \textbf{72.55} \\
    \textbf{F$^3$-Net ++} & 99.51  & 99.88  & 96.36  & 84.18  & 95.50  & 81.85  & 99.54  & 98.53  & 88.44  & \textbf{81.25} \\
    \textbf{LF ++} & 94.09  & 88.61  & 93.82  & 69.45  & 90.61  & 66.20  & 97.78  & 95.58  & 53.44  & \textbf{48.91} \\
    \textbf{NF ++} & 73.29  & 71.67  & 73.51  & 67.41  & 78.00  & 67.11  & 79.64  & 68.89  & 65.98  & \textbf{64.66} \\
    \textbf{Average} & 92.36  & 90.89  & 91.40  & 75.40  & 91.22  & 71.61  & 94.37  & 92.26  & 66.84  & \textbf{59.74} \\
    \bottomrule
    \end{tabular}%
  \label{tab_strong_def}%
\end{table}%

\begin{figure}[]
    \centering
    \includegraphics[width=0.47\textwidth]{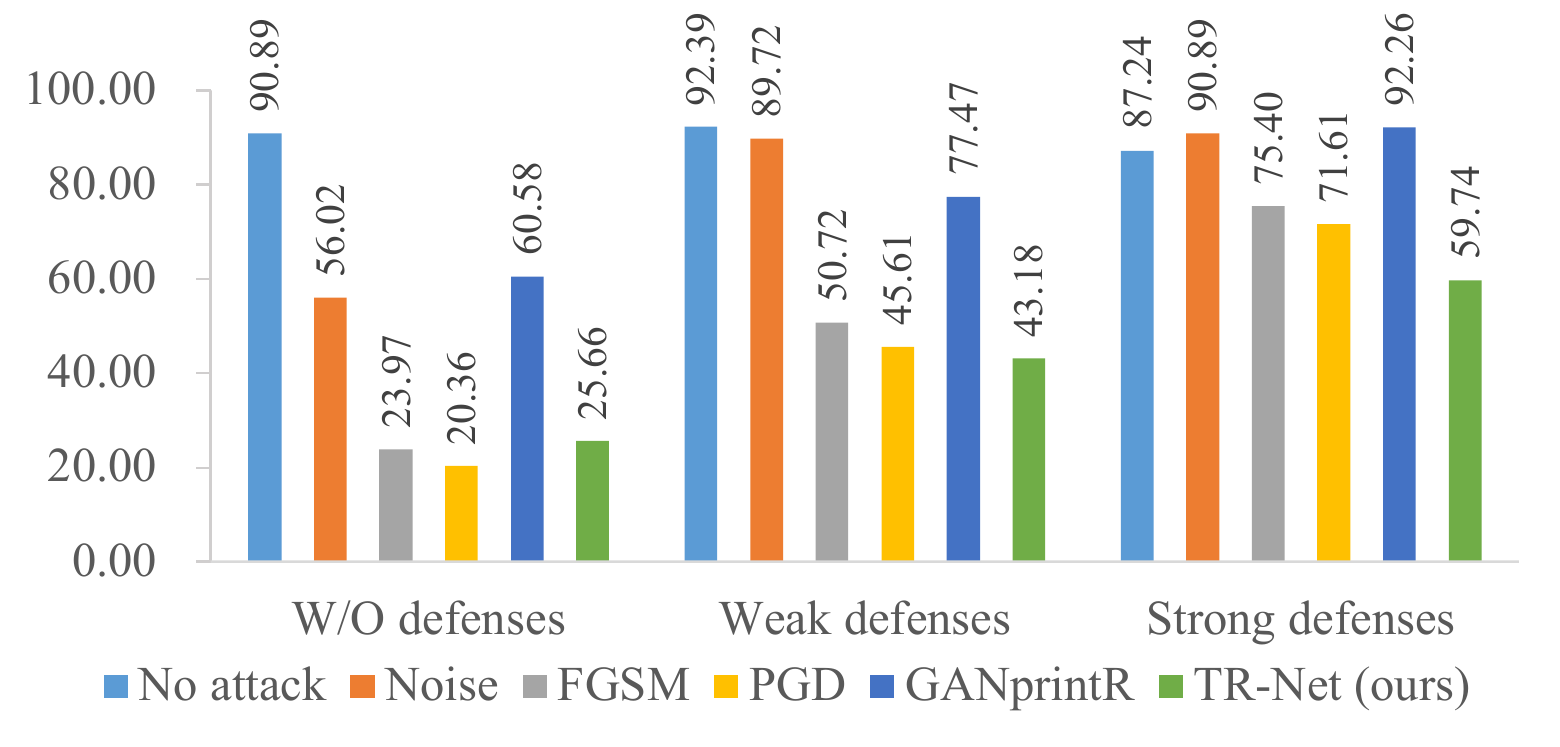}
    \caption{The average detection accuracy under the five attack methods in three defensive strategy groups.}
    \label{fig_avgacc}
\end{figure}

\paragraph{Discussion}
Figure \ref{fig_avgacc} offers a more intuitive comparison of the average detection accuracy given the five attack methods in the three defensive strategy groups. The trace removal attack was the most effectiveness in all groups. In the circumstances where detectors are defended with data augmentation strategies, especially the adversarial augmentation strategy, the baseline attack methods generally undergo a considerable loss of efficacy, while TR-Net continues to pose a threat, and the threat is even more serious than the white-box adversarial attacks in the two defensive groups (the red color results in \ref{tab_weak_def}-\ref{tab_strong_def}). In addition, as shown in Tables \ref{tab_wo_def}-\ref{tab_strong_def}, the TR-Net shows superior transferablity across different detectors compared to the baselines. Again, we emphasize that, unlike the baseline attacks, our attack was implemented upon all detectors being completely unknown during training. \textit{In conclusion, the attacking goal of fraudulence is well satisfied by the proposed trace removal attack.}

\subsubsection{Evaluating stealthiness}
As mentioned, this goal was evaluated based on visual quality. To be considered stealthy, a given attack sample was required to be perceptually indistinguishable from the corresponding DeepFake image.
    
\begin{figure}[]
    \centering
    \includegraphics[width=0.48\textwidth]{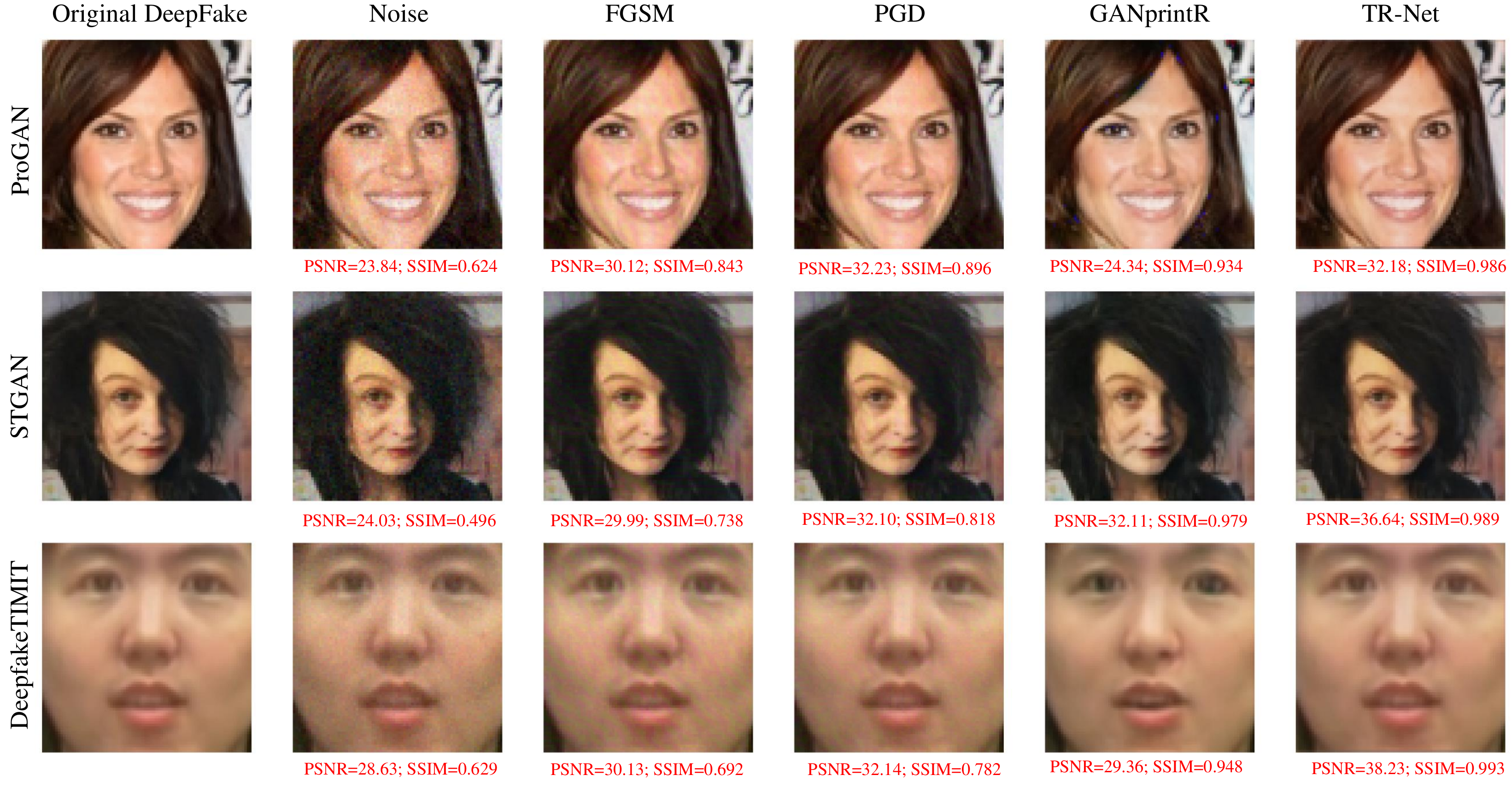}
    \caption{Examples of three types of DeepFake images and the corresponding attack samples from different attack methods.}
    \label{fig_visualquality}
\end{figure}

Table \ref{tab_visualquality} demonstrates the visual quality differences between DeepFake samples and attack samples in the evaluation set in terms of the average PSNR and SSIM scores. \textbf{TR-Net} achieved the highest PSNR ($35.16\pm3.32 db$) and SSIM ($0.988\pm0.004$) scores, indicating that attack samples generated by \textbf{TR-Net} contain less noise and have a visual quality closer to the original DeepFake samples. Figure \ref{fig_visualquality} also provides a qualitative view of the examples of three types of DeepFake images and the corresponding attack samples from different attack methods. As shown in the figure, the methods that add noise, including including \textbf{Noise}, \textbf{FGSM} and \textbf{PGD}, bring perceptual noise or a blurriness to the attack samples that may have be potentially screened out by the forensic investigators. By contrast, the reconstruction-based methods, especially \textbf{TR-Net}, generated high-fidelity attack samples that were perceptually similar to the original ones. \textit{Thus, we can conclude that the goal of stealthiness is well satisfied as well.}

\begin{table}[]
  \centering
  \scriptsize
  \caption{The visual quality comparison of different attack samples in terms of the average PSNR and SSIM scores. Each bold value indicates the best result in the corresponding line.}
  \renewcommand\arraystretch{1.2}
  \setlength{\tabcolsep}{0.6mm}{
    \begin{tabular}{l|ccccc}
    \toprule
            &\textbf{Noise} & \textbf{FGSM} & \textbf{PGD} & \textbf{GANprintR} & \textbf{TR-Net} (ours) \\
    \midrule
    PSNR (db) & 26.86$\pm$3.24  & 30.13$\pm$0.08 &	32.23$\pm$0.20 &	25.80$\pm$2.58 & \textbf{35.16}$\pm$3.32 \\
    SSIM & 0.634$\pm$0.149 &	0.764$\pm$0.059	& 0.836$\pm$0.041 &	0.924$\pm$0.050	& \textbf{0.988}$\pm$0.004  \\
    \bottomrule
    \end{tabular}}%
  \label{tab_visualquality}%
\end{table}%

\subsection{Attacking with limited background knowledge}
Our next set of scenarios imposes restrictions on the attacker’s background knowledge of data. Here, only the \textbf{Xception} and \textbf{F$^3$-Net} detectors are retained since the two have shown a relatively better ability in detecting DeepFakes. The average PSNR and SSIM scores are also reported.

\subsubsection{Limited dataset size}
To simulate that the attacker only has access to a limited dataset, we randomly sampled six subsets from the full training set, containing $1\%$, $5\%$, $10\%$, $25\%$, $50\%$, $75\%$ and $100\%$ of data (i.e., for $1\%$ that equates to a total of $660$ semantically-closest pairs of real and fake images). Then we trained TR-Net from scratch on each subset individually and evaluated our results on the same evaluation dataset as in the previous scenario.

\begin{figure}
    \centering
    \includegraphics[width=0.5\textwidth]{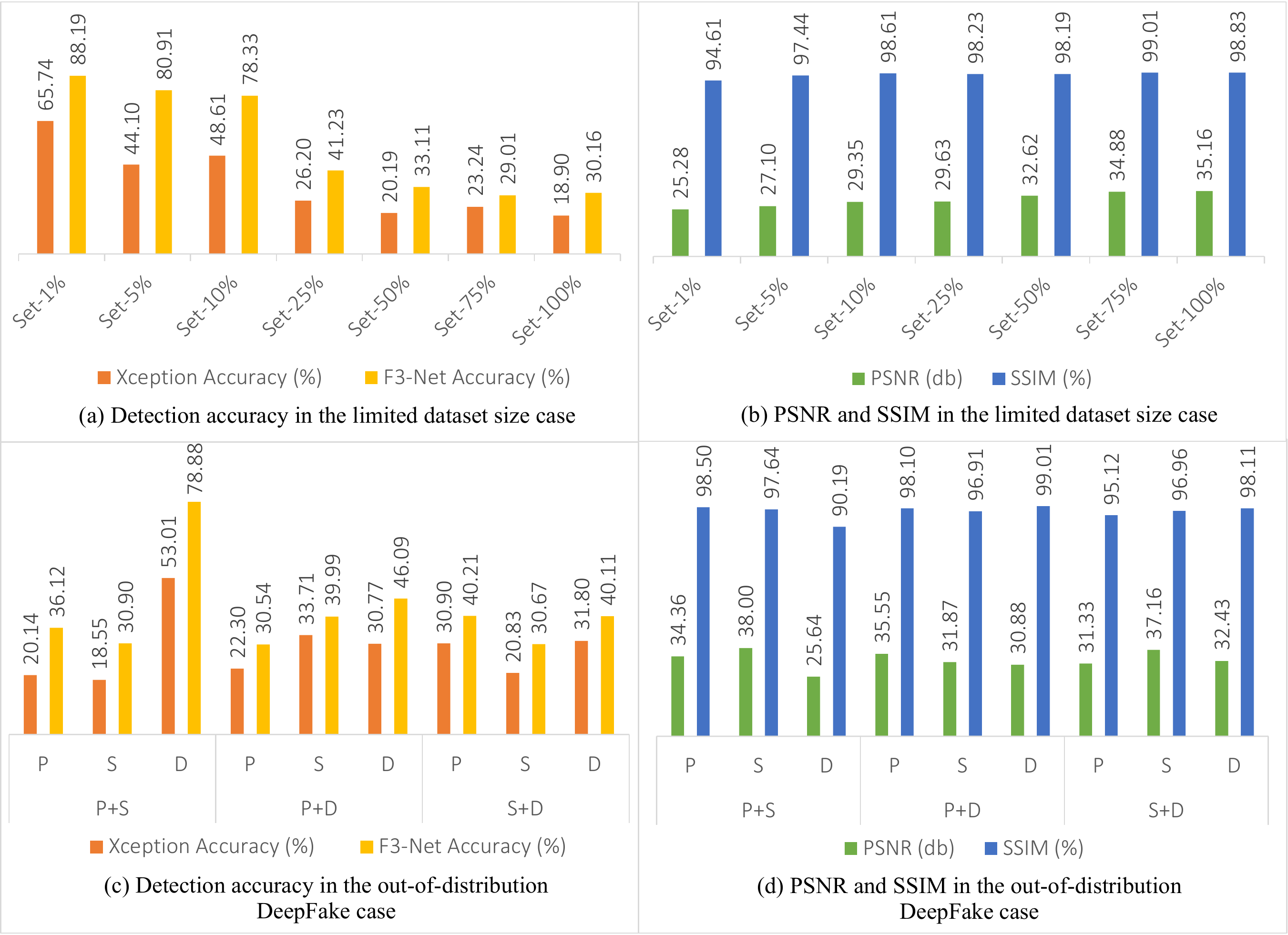}
    \caption{Attacking performance in the settings where the attacker is imposed with different restrictions on data accessibility. P: ProGAN; S: STGAN; D: DeepfakeTIMIT}
    \label{fig:limit_knowledge}
\end{figure}

Figure \ref{fig:limit_knowledge}.a and \ref{fig:limit_knowledge}.b illustrates the detection accuracy and PSNR and SSIM scores for each subset. From the results, it appears there is a threshold for the dataset size that is within $10\%-25\%$, under which both the accuracy and visual quality are affected. This is unsurprising since attack methods based on GAN learning are essentially data-driven. However, when the training set size equates to more than a quarter of the original data set, all metrics increase rapidly and remain relatively stable at a satisfactory level. The results indicate that TR-Net fits well even with a relatively small amount of training samples which are easily collected. This weak data volume-dependency makes TR-Net practically feasible.

\subsubsection{Out-of-distribution DeepFake}
We also assessed TR-Net’s performance on out-of-distribution DeepFakes to demonstrate its domain-independency. In this scenario, the attacker was restricted to train the model with only two types of DeepFake images. Yet the evaluation set still contained all three DeepFake types. Here, for example, "P+S" (short for “ProGAN+STGAN”) indicates training with ProGAN and STGAN samples.

Figure \ref{fig:limit_knowledge}.c and \ref{fig:limit_knowledge}.d show the results for detection accuracy and visual quality when trained with different training groups. What is shown is that, when implementing the attack on the samples generated by an unknown DeepFake method that is not included in the training set, the resulting attack samples suffer from a decrease in both detection-evasive ability and visual quality. For instance, comparing the ProGAN results in the ``P+S" group (where ProGAN is included in the training set) with those in the ``S+D" group (where ProGAN is not included in the training set), the \textbf{Xception}'s detection accuracy increases from $36.12\%$ to $40.21\%$ and the \textbf{F$^3$-Net}'s detection accuracy increases from $20.14\%$ to $30.90\%$ (Figure \ref{fig:limit_knowledge}.c). Also, the PSNR scores decreased from $34.36 db$ to $31.33 db$ and the SSIM scores decreased from $98.50\%$ to $95.12\%$ (Figure \ref{fig:limit_knowledge}.d). The performance degradation was much more significant for unknown DeepfakeTIMIT samples than that for unknown ProGAN and STGAN samples. The reason is that both the source ProGAN and STGAN models were pretrained with the CelebA dataset while the source DeepfakeTIMIT model are developed with another dataset, where there exists a domain inconsistency. Our findings suggest that fine-tuning TR-Net in a domain to be consistent with the target detector helps to improve the efficacy of the attack.

\subsection{A closer look into trace removal}
In this section, we offer a closer look at the trace removal to justify the DeepFake trace discovery outlined in Section \ref{sec:DTD}. This examination helps us to understand why and how TR-Net removes all traces.
\begin{figure}[]
    \centering
    \includegraphics[width=0.48\textwidth]{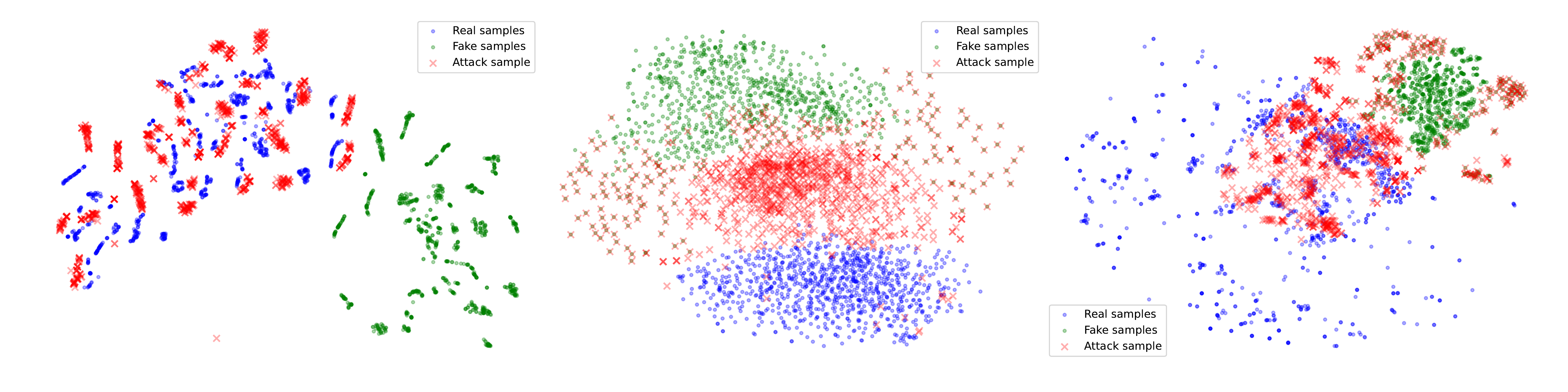}
    \caption{Trace features in the latent spaces learned by, from left to right, the spatial discriminator $D_{1}$, the spectral discriminator $D_{2}$ and the fingerprint discriminator $D_{3}$. t-SNE is used to project the representations of features from each discriminator’s last convolutional layer onto its two principal components.}
    \label{fig_tsne}
\end{figure}

\begin{figure*}[]
	\centering
	\subfigure[ProGAN group]{
		\begin{minipage}[b]{0.3\textwidth}
			\includegraphics[width=1\textwidth]{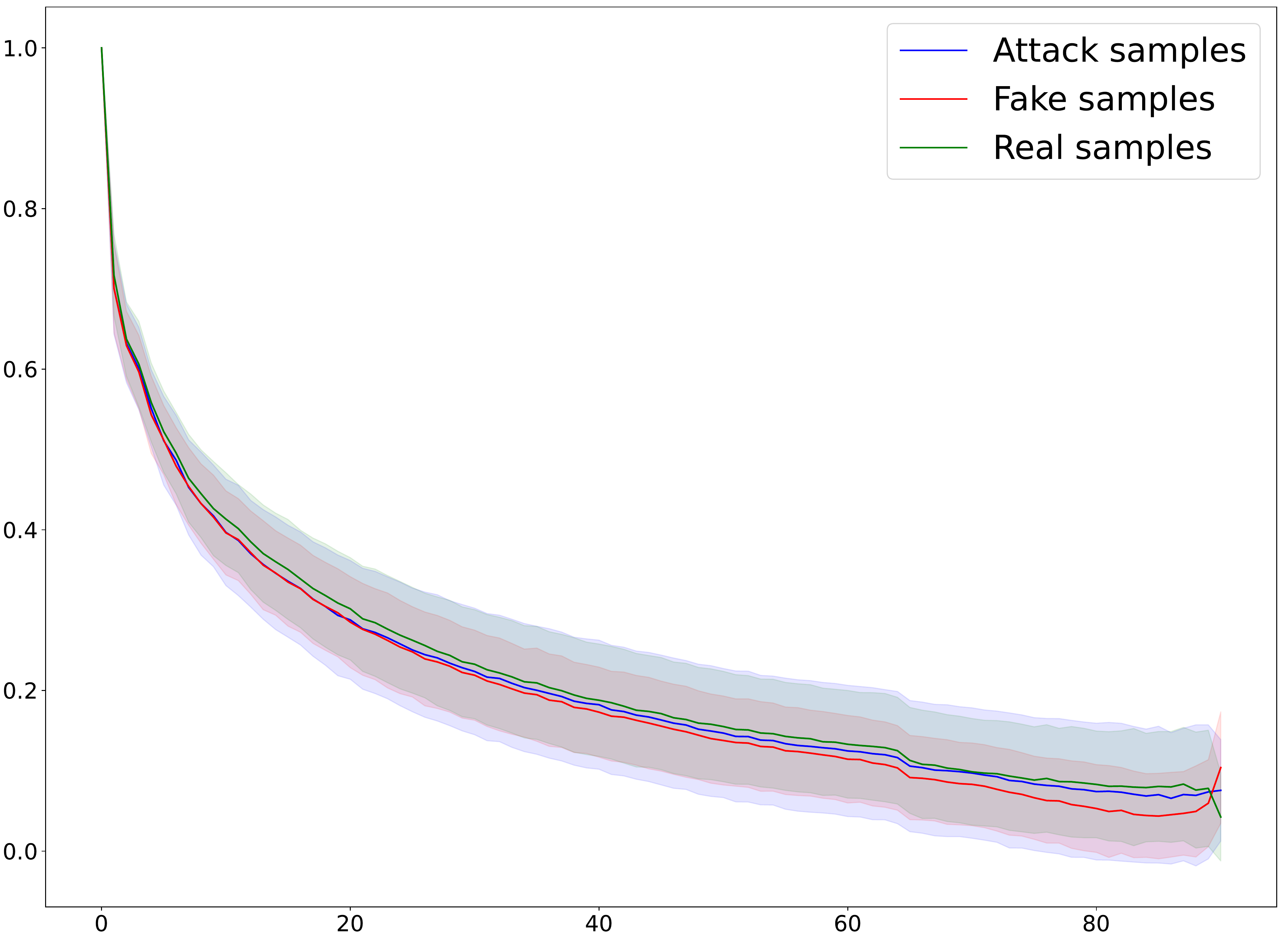} 
		\end{minipage}
		\label{frequency_dis_progan}
	}
    	\subfigure[STGAN group]{
    		\begin{minipage}[b]{0.3\textwidth}
   		 	\includegraphics[width=1\textwidth]{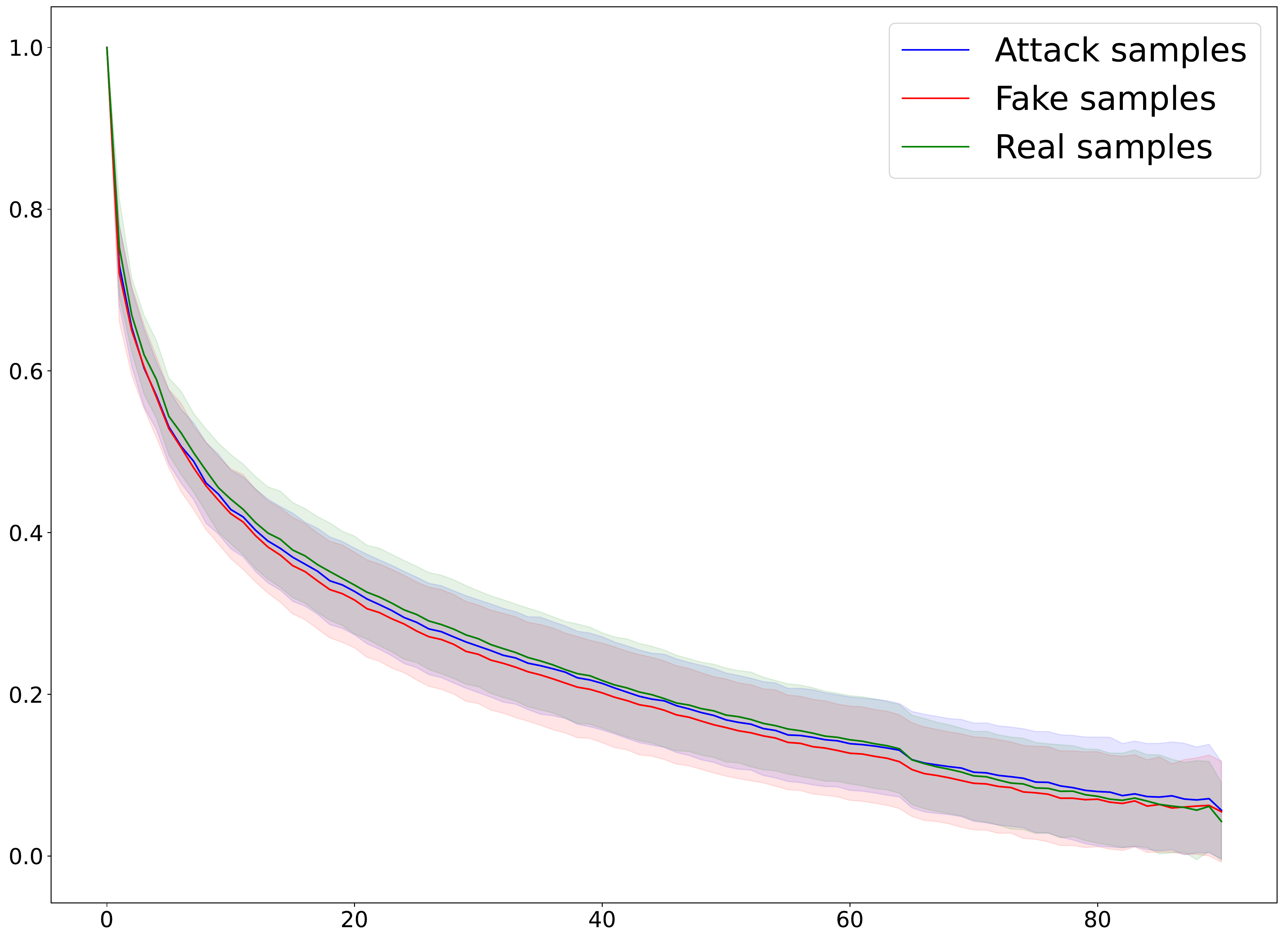}
    		\end{minipage}
		\label{frequency_dis_stgan}
    	}
	\subfigure[DeepfakeTIMIT group]{
		\begin{minipage}[b]{0.3\textwidth}
			\includegraphics[width=1\textwidth]{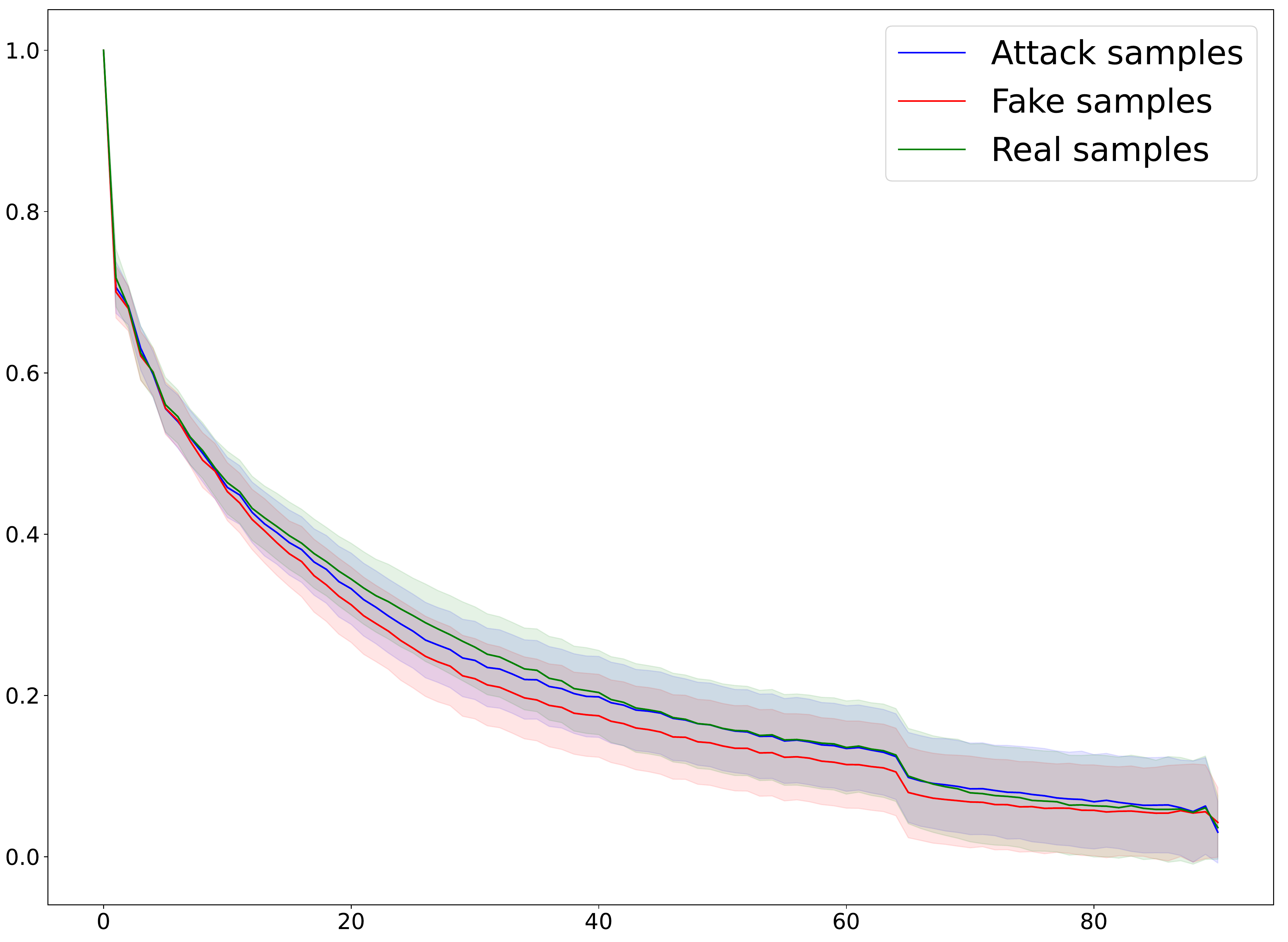} 
		\end{minipage}
		\label{frequency_dis_deepfake}
	}
	\caption{The power spectral density distributions of real, DeepFake and attack samples for three DeepFake types. Zooming in for a better visualization.}
	\label{fig:frequency_dis}
\end{figure*}

\begin{figure}[]
    \centering
    \includegraphics[width=0.3\textwidth]{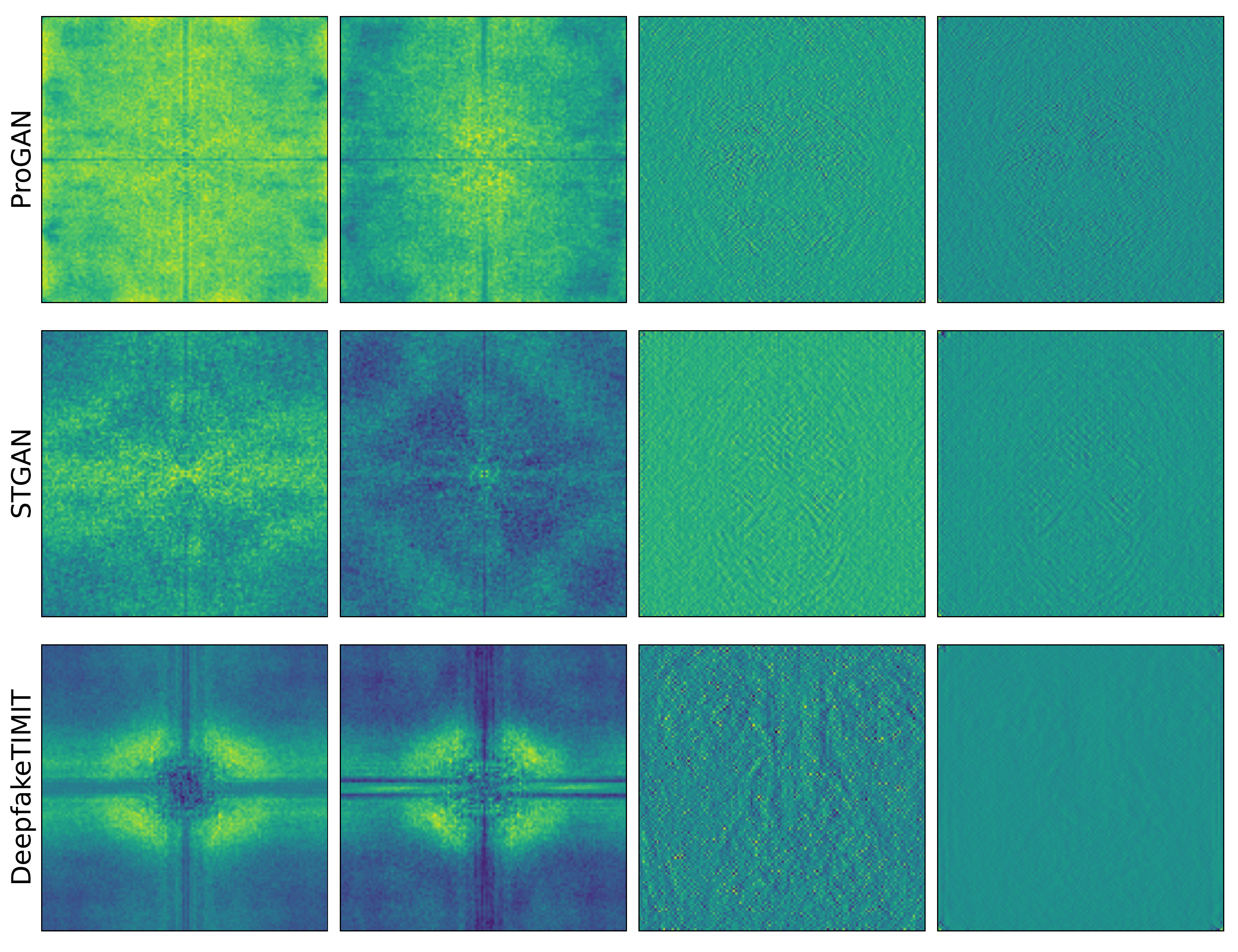}
    \caption{The average spectra and noise residual differences for three DeepFake types. The first two columns are spectrum differences between real and source fake samples and between real and attack samples, respectively; The last two columns are noise residual differences between real and source fake samples and between real and attack samples, respectively. A brighter entry means a bigger difference.}
    \label{fig_differences}
\end{figure}
\subsubsection{Explanation in the feature space}
The representations of different DeepFake traces are well learned by a set of discriminators, thus, we analyzed the geometrical shifting of trace features encoded by each discriminator in the latent feature space. Since that the generator and discriminators are trained in parallel, the discriminators $\mathbb{D}^{*}$ resulting from the same checkpoints of the optimal generator $G^{*}$ are adopted as the trace feature descriptors. The trace features output by the last $512*4*4$ convolutional layer of $\mathbb{D}^{*}$ are analyzed. t-SNE \cite{van2008visualizing} is performed to reduce feature dimensionality, so as to obtain an interpretable two-dimensional view of geometrical shifting.  

The result from each discriminator is shown individually in Figure \ref{fig_tsne}. Each discriminator corresponds to a single trace type. We can see that there is a distinct trend that the attack samples’ trace features are transferring towards the the real images’ features. The result confirms our conjecture that TR-Net can reduce the distribution gap between the attack samples and the real samples at the trace feature level via adversarial learning. In addition, the spectral traces from $D_{1}$ and the fingerprint traces from $D_{3}$ have a more significant migration than the spectral traces from $D_{2}$. This occurs because of the aforementioned optimization conflict between the semantic features and the trace features in latent space. Since the frequency components are closely correlated with both the trace and semantic information where no distinct boundary applies, weakening the trace representations of the DeepFake samples while retaining their visual information must lead to a sub-optimum. Even so, the attack efficacy is barely affected as shown in previous experiments.

\subsubsection{Explanation in the frequency and noise spaces}
Then, we explain the trace removal in the frequency and noise space to further justify the trace removal success. First, we compare the PSD distributions of real, DeepFake and attack samples. Figure \ref{fig:frequency_dis} shows the average PSD distribution along with standard deviation for different DeepFake types. The results show that for all DeepFake types, the distribution gaps between attack and real samples are significantly smaller than those between attack and fake samples. The result again showcases the wide applicability of the trace removal attack to various DeepFake types. Meanwhile, the gaps between attack and real samples are slighter in the STGAN and DeepfakeTIMIT groups than in the ProGAN group. We suppose the reason is associated with the semantically-closest pairs. For the semantically-closest pairs in both the STGAN and DeepfakeTIMIT groups, each fake sample has an exact source real image as a counterpart. In contrast, the fake samples in the ProGAN’s semantically-closest pairs correspond to their nearest-neighbor similar real images where a larger visual difference exists, leading to underfitting in the frequency domain. 
 
We also provide some qualitative results as complementary evidence. Figure \ref{fig_differences} illustrates the average spectra and noise residual differences between the real and source fake samples and between the real and attack samples. A brighter entry indicates a larger difference. As shown in the figure, the differences between the real and source fake samples are much more significant than those between the real samples and the attack samples. This result further highlights the fact that successful trace removal will refine the DeepFake images to be closer to the real ones, which they can deceive arbitrary detectors.

\section{Conclusion and future work}
In this paper, we focused on an anti-forensics attack against DeepFake detectors. We presented a novel detector-agnostic attack, called a trace removal attack, that is capable of refining DeepFake images by removing all possible DeepFake traces via an one-versus-multiple adversarial learning network. The refined DeepFake images are closer to the real images and can therefore bypass arbitrary and even unknown detectors. We assessed the efficacy of the trace removal attack against a wide range of state-of-the-art detectors in heterogeneous high-level security scenarios where the detectors were embedded with various defensive strategies and the attacker’s knowledge of data was limited. Our findings reveal that, the proposed trace removal attack achieves the highest attack effectiveness while introducing minimal visual quality loss compared with contemporary adversarial and reconstruction-based attacks. In the future, we will focus on developing more robust forensics countermeasures against trace removal attacks.

\bibliographystyle{IEEEtran}
\bibliography{IEEEabrv,./reference.bib}

\end{document}